\documentclass[runningheads]{llncs}
\usepackage{graphicx}

\usepackage{tikz}
\usepackage{comment}
\usepackage{amsmath,amssymb} 

\usepackage{hyperref}
\hypersetup{
    urlcolor=cyan,
    }
\usepackage{color}
\usepackage{comment}
\usepackage{multirow}

\let\oldnorm\norm
\def\norm{\@ifstar{\oldnorm}{\oldnorm*}}
\makeatother
\usepackage[accsupp]{axessibility}  

\usepackage{pifont}
\usepackage[capitalize]{cleveref}
\usepackage[normalem]{ulem}
\usepackage{mathrsfs}
\usepackage{array}
\usepackage{mathrsfs}
\usepackage{array}
\makeatletter
\newcommand{\thickhline}{%
    \noalign {\ifnum 0=`}\fi \hrule height 1pt
    \futurelet \reserved@a \@xhline
}
\usepackage[capitalize]{cleveref}

\crefname{section}{Sec.}{Secs.}
\Crefname{section}{Section}{Sections}
\Crefname{table}{Table}{Tables}
\crefname{table}{Tab.}{Tabs.}

\newcommand{\eg}{\textit{e}.\textit{g}.}
\newcommand{\etal}{\textit{et al.}}
\begin{document}
\pagestyle{headings}
\mainmatter
\def\ECCVSubNumber{3601}  

\title{Event-guided Deblurring of Unknown Exposure Time Videos} 

\titlerunning{Event-guided Deblurring of Unknown Exposure Time Videos}
%
\author{Taewoo Kim\inst{1} \and
Jeongmin Lee\inst{1} \and
Lin Wang \inst{2} \and Kuk-jin Yoon\inst{1}}
\authorrunning{Kim et al.}
%
\institute{$^1$ Korea Advanced Institute of Science and Technology \\
\email{\{intelpro, jeanmichel, kjyoon\}@kaist.ac.kr}\\
$^2$AI Thrust, HKUST Guangzhou and Dept. of CSE, HKUST \\
\email{linwang@ust.hk}}
\maketitle
\begin{abstract}
Motion deblurring is a highly ill-posed problem due to the loss of motion information in the blur degradation process. Since event cameras can capture apparent motion with a high temporal resolution, several attempts have explored the potential of events for guiding deblurring. These methods generally assume that the exposure time is the same as the reciprocal of the video frame rate. However, this is not true in real situations, and the exposure time might be unknown and dynamically varies depending on the video shooting environment(e.g., illumination condition). In this paper, we address the event-guided motion deblurring assuming dynamically variable unknown exposure time of the frame-based
camera. To this end, we first derive a new formulation for event-guided motion deblurring by considering the exposure and readout time in the video frame acquisition process. We then propose a novel end-to-end learning framework for event-guided motion deblurring. In particular, we design a novel Exposure Time-based Event Selection(ETES) module to selectively use event features by estimating the cross-modal correlation between the features from blurred frames and the events. Moreover, we propose a feature fusion module to fuse the selected features from events and blur frames effectively. We conduct extensive experiments on various datasets and demonstrate that our method achieves state-of-the-art performance. 
Our project code and dataset are available at: \url{https://intelpro.github.io/UEVD/}
\end{abstract}

\section{Introduction}

Motion blur often occurs due to the non-negligible exposure time of the frame-based cameras. Any motion during the video recording makes the sensor observe an averaged signal from different points in the scene~\cite{wieschollek2017learning,noroozi2017motion}. Motion deblurring is a task aiming at restoring sharp frame from the motion-blurred ones. This task is a highly ill-posed problem due to the loss of motion information in the blur degradation process, especially in the complex real-world scene~\cite{Gao_2019_CVPR,argaw2021motion,jiang2020learning}. 
Recently, deep learning (DL)-based approaches have achieved great success in modeling general motion blur and recovering sharp frames from the motion-blurred frames~\cite{zhang2019deep,nah2017deep,shen2020blurry,zhang2020video}. However, they are limited to specific scenarios and may fail to recover the sharp frames for the severe motion blur. 
Event cameras are bio-inspired sensors that encode the per-pixel intensity change asynchronously with high temporal resolution.

\begin{figure}[!t]
\centering
\includegraphics[width=0.9\columnwidth]{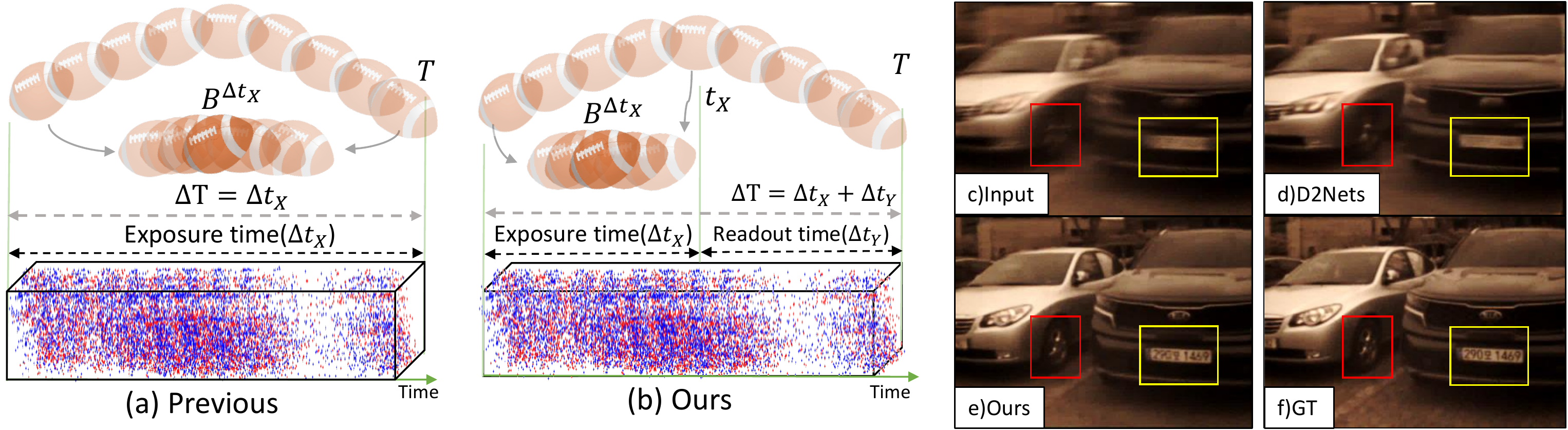}
\caption{(a-b)The motivation of our work. The figure shows the image formation setting assumed in (a) previous event-guided methods and (b) our work. Here, $\Delta T$ represents the shutter period. From above to bottom, continuous latent images, video frame acquisition, corresponding event streams($E^{\Delta T}$), respectively. Previous methods generally assume that the exposure time is the same as the reciprocal of the video frame rate, while ours considers the actual exposure time for selectively using events. (c-f) Results of deblurring on the unknown exposure time video frame. Our method restores a sharper frame than the existing event-guided deblurring methods, \eg, D2Nets~\cite{Shang_2021_ICCV}, trained on the same dataset.}
\label{fig1:video_frame_acquisition}
\end{figure}

Many endeavors have been engaged in reconstructing image/video from event streams~\cite{rebecq2019high,wang2020event,mostafaviisfahani2021e2sri,Wang_2020_CVPR,zou2021learning,wang2021evdistill,wang2021dual}. However, the reconstructed results from the events may lose texture details. 
Consequently, several attempts have leveraged events for guiding motion deblurring~\cite{lin2020learning,jiang2020learning,zhang2021fine,pan2019bringing,wang2020event,Shang_2021_ICCV}, trying to take advantage of frame-based and event cameras.
As shown in Fig.~\ref{fig1:video_frame_acquisition}(a), these methods generally assume that the exposure time is the same as the reciprocal of the video frame rate and perform the deblurring guided by the events within the exposure time. 
However, this assumption may not be valid in real situations as the video frame acquisition process generally consists of two phases: exposure phase ($X$) and readout phase ($Y$)~\cite{internet_video_frame_acquision,zhang2020video}, as shown in Fig.~\ref{fig1:video_frame_acquisition}(b).
In the exposure phase, the camera's shutter opens and receives lights.
In the readout phase, the camera clears charge from the serial register, and the pixel value is digitalized.
The total time, reciprocal of the frame rate, is called the shutter period, not exposure time. 
Since the motion blur of the frame-based camera occurs only in the exposure phase rather than the readout phase, it is crucial to use the events during the exposure phase within the shutter period.
However, the exposure time is not always known, and furthermore, it can dynamically vary depending on the imaging environments when the auto-exposure function turns on.

For that reason, we assume the exposure time is unknown when performing the event-guided motion deblurring, as shown in Fig.~\ref{fig1:video_frame_acquisition}(b), to consider more practical situations.
This assumption can lead to significant performance changes. 
If we apply the existing event-guided deblurring methods, \eg,~\cite{Shang_2021_ICCV}, for unknown exposure time video frames, the performance degrades, as shown in  Fig.~\ref{fig1:video_frame_acquisition}(d).\footnote{This result is obtained by using all the events during the shutter period without knowing the actual exposure time.} Therefore, it is necessary to infer the actual exposure time to use events rightly for deblurring. Accordingly, we propose an end-to-end learning framework.
As the exposure time is assumed to be unknown, we first propose an event selection module called the Exposure Time-based Event Selection(ETES).
The proposed module extracts the relevant events within the shutter period by estimating the temporal correlation between events and blur frames features.
As such, only the event features corresponding to the exposure time are automatically selected for guidance.
Second, we propose a new module for events-frame feature fusion. Such a fusion module leads to more robust feature representation learning.
Lastly, as a lack of publically available real-world event datasets for event-guided motion deblurring, we collect color images and event data in diverse real-world scenes using a DAVIS-346 color event camera.
We then make a dataset by simulating dynamically variable exposure time using collected frames and real events for performing motion deblurring without exposure time information.

In summary, our contributions are four-fold.
(I)  We study and formulate an event-guided deblurring for unknown exposure time videos.
(II) Based on the formulation, we design a novel event feature selection method within exposure time and propose a feature fusion module to use complementary information of events and frame features.
(III) We build a novel large-scale dataset for event-guided motion deblurring, including RGB images and the real events in various scenes. 
(IV) We conduct various experiments on the synthetic event and our real-world event datasets and demonstrate that our method achieves new state-of-the-art performance.

\section{Related Works}
\noindent\textbf{Image and Video Deblurring.} 
DL has been broadly applied to image and video deblurring. 
Earlier works, \eg, \cite{su2017deep}, utilized convolutional neural networks (CNNs) with frame alignment and merged multiple frames based on the homography for video deblurring.
The baseline networks have been improved by applying more sophisticated network structures or learning methods, \eg, recurrent neural networks (RNN)~\cite{wieschollek2017learning,hyun2017online,tao2018scale,nah2019recurrent,zhong2020efficient,park2020multi}, multi-scale architecture~\cite{nah2017deep,Brehm_2020_CVPR_Workshops,cho2021rethinking}, adversarial training~\cite{zhang2018adversarial,Kupyn_2018_CVPR,Kupyn_2019_ICCV,zhang2020deblurring,Zhang_2020_CVPR}, multi-stage approaches~\cite{zhang2019deep,zamir2021multi,chen2021hinet}, video frame alignment~\cite{kim2018spatio,Zhou_2019_ICCV,Wang_2019_CVPR_Workshops,Pan_2020_CVPR,li2021arvo} in an end-to-end learning manner. \\
\noindent\textbf{Event-guided Motion Deblurring.}\; The event cameras show higher temporal resolution and HDR properties. 
To leverage the advantages of event cameras,  recent works focus on event-guided deblurring.
Pan \etal~\cite{pan2019bringing,pan2020high} first proposed a deblurring framework by formulating an event-based double integral model(EDI).
Although they show the effectiveness of formulation, they often fail to reconstruct the scene details due to the noisy contrast threshold of the events. \
To solve the aforementioned issues, Jiang \etal~\cite{jiang2020learning} introduced a DL-based deblurring framework by using an RNN-based network architecture and a directional event filtering module.
More recently, Lin \etal~\cite{lin2020learning} proposed a CNN-based framework driven by an event-based physical model for deblurring and frame interpolation.
Shang \etal~\cite{Shang_2021_ICCV} proposed an event-guided deblurring framework to exploit the non-consecutively blurry video frames.
Concurrently, Xu \etal~\cite{Xu_2021_ICCV} proposed a self-supervised learning framework that utilizes real events to alleviate performance degradation due to the domain gap between real and synthetic data.
These works generally assume the exposure time is the same as the shutter period. However, as aforementioned, this assumption is not valid in many real situations.
\textit{Unlike these works, we propose a novel framework for unknown exposure time videos.}

\section{Method}
\subsection{Formulation}\label{formulation}
\noindent\textbf{Event Selection}\;
A video frame acquisition consists of the exposure phase $X$ and readout phase $Y$, as depicted in Fig.~\ref{fig1:video_frame_acquisition}. We denote the duration of the exposure phase as $\Delta t_{X}$ and the readout phase as $ \Delta t_{Y}$. 
The summation of two phases (shutter period), $\Delta T$, represents the time to acquire one video frame.
By the nature of frame-based cameras, motion blur only occurs during the exposure phase $X$.
In contrast, events are generated in the exposure phase $X$ and the readout phase $Y$.
Therefore, it is imperative to use events during the exact duration of the exposure phase $\Delta t_{X}$ only.
The existing event-guided deblurring methods~\cite{pan2019bringing,lin2020learning} generally assume that the exposure time is equal to the shutter period; $\Delta T=\Delta t_{X}$. By contrast, our goal is to estimate the temporal correlation between the motion-blurred frame and the event during $\Delta T$ to handle the unknown exposure time $\Delta t_{X}$.
A motion-blurred frame can be expressed as the temporal average of $N$ latent frames during the exposure time $\Delta t_{X}$ as
\begin{gather}
B^{\Delta t_{X}}(x,y) 
\simeq \frac{1}{N}{\sum}_{i=1}^{N}L_{\tau_{i}}(x,y), \label{eq:blur_formulation}
\end{gather}
where $B^{\Delta t_{X}}(x,y)$ denotes a blurred frame, and $L_{\tau_{i}}(x,y)$ is the $i$-th latent frame at $\tau_{i}$. 
For event cameras, an event is generated when the log intensity change exceeds a contrast threshold $\beta$.
\begin{equation}
   E^{t}(x,y) = \bigg\{\begin{matrix} 
\! +1, \quad if \ \log(\frac{I^t (x,y)}{I^{t-1}(x,y)}) \geq \beta \\ 
-1,\quad if \ \log(\frac{I^t (x,y)}{I^{t-1}(x,y)}) \leq -\beta 
   \end{matrix} 
\end{equation}
where $I^t(x,y)$ is the intensity value at timestamp $t$.
 Given two consecutive frames, $I^{t_{1}}(x,y)$ and $I^{t_{2}}(x,y)$, the events $E^{t}(x,y)$ are generated by intensity changes between them. Accordingly, we can derive the relationship between two intensity images based on event generation.
\begin{equation}
\begin{split}
I^{t_{2}}(x,y) \simeq I^{t_{1}}(x,y)\cdot \mathrm{exp}( {\sum}_{t_{1}}^{t_{2}}  \beta \cdot E^{t}(x,y)) = I^{t_{1}}(x,y)\tilde{R}(x,y)\label{eq:event_interpolation}
\end{split}
\end{equation}
By combining Eq.(\ref{eq:blur_formulation}) and Eq.(\ref{eq:event_interpolation}), we can represent a blurred frame as follows:
\begin{gather}
B^{\Delta t_{X}}(x,y) \simeq I^{t}(x,y)(\frac{1}{N}{\sum}_{i=1}^{N}\tilde{R}_{i}(x,y)) = I^{t}(x,y)S(x,y)\label{eq:blur_with_event_eqn}
\end{gather}
where $I^{t}(x,y)$ denotes a latent frame, which is the result of deblurring, and $S(x,y)$ is the summation of residual matrix $\tilde{R}_{i}(x,y)$.
According to Eq.(\ref{eq:blur_with_event_eqn}), we need to estimate a $S(x,y)$ corresponding to the exposure time of the blurred frame. However, since the exact exposure time is assumed to be unknown, we aim at estimating the temporal correlation between the blurred frames and events during $\Delta T$.
That is, to estimate $S(x,y)$, we find a function $f_{\theta^{*}}$ that approximates a set of events $\{E^{\Delta t_{X}}\}$ based on $\{E^{\Delta T}\}$ and the blurred frame $B^{\Delta t_{X}}$ as
\begin{gather}
    f_{\theta^{*}} \sim \{ E^t \mid \psi(E^t, B^{\Delta t_{X}})>0 \}; \; \text{ } t \in \{0, T\}
    \label{eq:event_selection}
\end{gather}
where $\psi(E^{t}, B^{\Delta t_{X}})$ is a conditional function obtained by calculating the temporal correlation between $E^{t}$ and $B^{\Delta t}$.
With respect to $E^{t}$, a function $\psi(E^{t}, B^{\Delta t_{X}}) $ is true when the intensity change corresponding to $E^{t}$ exists in the motion-blurred frame $B^{\Delta t_{X}}$ and vice versa.
As such, only the events within the exposure time are selected as guidance.
\begin{figure}[!t]
\centering
\includegraphics[width=1.0\linewidth]{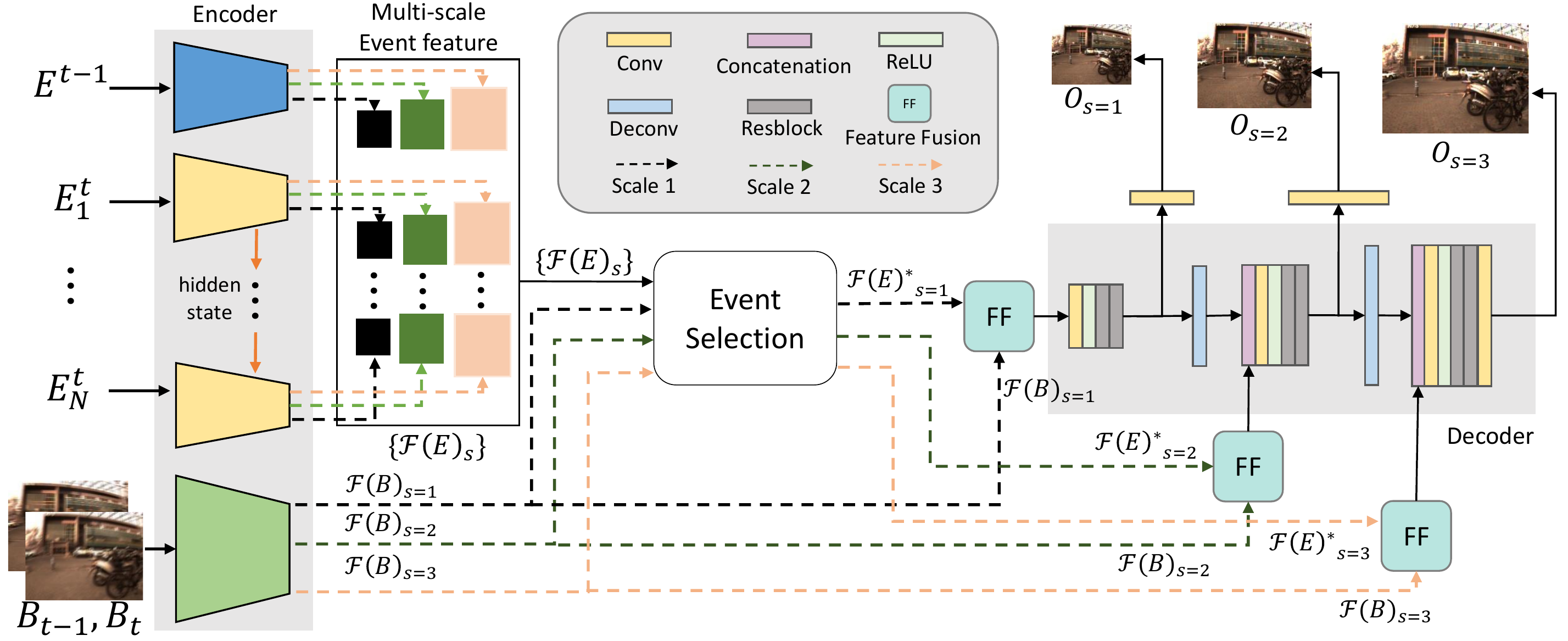} 
\caption{Overview of the proposed framework. For the encoder, blue, yellow, and green boxes represent an event encoder for the past part, a shared RNN-based event encoder for the current part, and a blur-frame encoder, respectively.}
\label{fig2:overall_network_architecture}
\end{figure}

\noindent\textbf{Event-guided deblurring}\;
Through the above formulation, we can select event streams $\{E^{\Delta t_{X}}\}$ among $\{E^{\Delta T}\}$ for the unknown exposure time $\Delta t_X$. 
Accordingly, we aim at recovering an intermediate sharp video frame from the motion-blurred ones $B^{\Delta t_{X}}$ using $\{E^{\Delta t_{X}}\}$.
\subsection{The Proposed Framework}
\noindent Based on the formulation, we propose a novel end-to-end learning framework. To feed event streams to DNNs, we need to embed them to the fixed size tensor-like format. 
The voxel grid~\cite{zhu2019unsupervised} is a well-designed event representation as it preserves the spatio-temporal information of events.
We use 16 temporal bins of the voxel grid for all the experiments. \\
\noindent\textbf{Overview}
An overview of the proposed framework is illustrated in Fig.~\ref{fig2:overall_network_architecture}.
The network uses the past and current inputs together to reinforce the spatio-temporal dependency of the videos.
The overall framework consists of \textit{two} major components: \textit{event selection} and \textit{feature fusion}. First, we encode the embedded events via an RNN-based encoding network for event feature selection. Then, we propose a novel ETES module to select the event features corresponding to the unknown exposure time \textit{without any supervision}.
Second, in Sec.~\ref{fea_fuse_sec}, we propose a new feature fusion module that efficiently exploits the complementary information of selected events and frames.
After the two steps, our network processed fused feature in a coarse-to-fine manner using a pyramid structure.
\subsection{Event Selection}\label{event_selection_method}
\noindent \textbf{Recurrent Encoding for Embedded Events.}
To extract features from events, 
recent works widely adopt 2D CNNs~\cite{zhu2019unsupervised,lin2020learning}; however, they are less effective in preserving temporal information for the event selection under unknown exposure time(Eq.\ref{eq:event_selection}). 
Specifically, we first divide the events into the past shutter period part $\{E^{t-1}\}$ and the current shutter period part $\{E^{t}\}$.
We then use the current shutter-period events to infer unknown exposure time, and past information is only used for better deblurring.
Therefore, we use different event encoders
to allocate more channels to the current events for correctly
estimating exposure time.
For the past part $\{E^{t-1}\}$, we extract a feature pyramid $\{\mathcal{F}(E^{t-1})_{s}\} \in \mathbb{R}^{C_{s}^{U} \times H_{s} \times W_{s}}$, using a 2D CNN block with the scale index $s\in\{1,2,3\}$.
Especially for the current part $\{E^{t}\}$, we devise a new RNN-cell based event encoder with shared weights to extract features for preserving temporal information inspired by recent frame-based video deblurring works~\cite{zhong2020efficient,nah2019recurrent} as shown in Fig.~\ref{fig2:overall_network_architecture}.
Considering the temporal information of the current part, we first divide the voxel grid into $N$ temporal units with an equal time interval $\Delta t_{unit}$.
Thus, we get temporally divided event units $E^{t}_{n}\in\mathbb{R}^{2\times H \times W}$ with the temporal index $n \in \{1, ..., N\}$.
We then recursively update the hidden state of the RNN cell to reinforce temporal coherence between consecutive event units.
As such, we generate $N$ hierarchical feature maps at each scale index $s$ for the current part $\{\mathcal{F}(E^{t}_{1})_{s}, ...,  \mathcal{F}(E^{t}_{N})_{s}\}$.
The extracted features of the current and past parts are concatenated to form a feature pyramid, denoted by $\{\mathcal{F}(E)_{s}\} \in \mathbb{R}^{(N+1)\times C_{s}^{U} \times H_{s} \times W_{s}}$, where $C_{s}^{U}, H_{s}, W_{s}$ denotes the numbers of unit channels, height, and width at scale index $s$, respectively.
For brevity, we denote $N+1$ as $T$. 
The detailed network structures and RNN-based encoding methods are given in the supple. material.

\begin{figure}[!t]
\centering
\includegraphics[width=0.93\columnwidth]{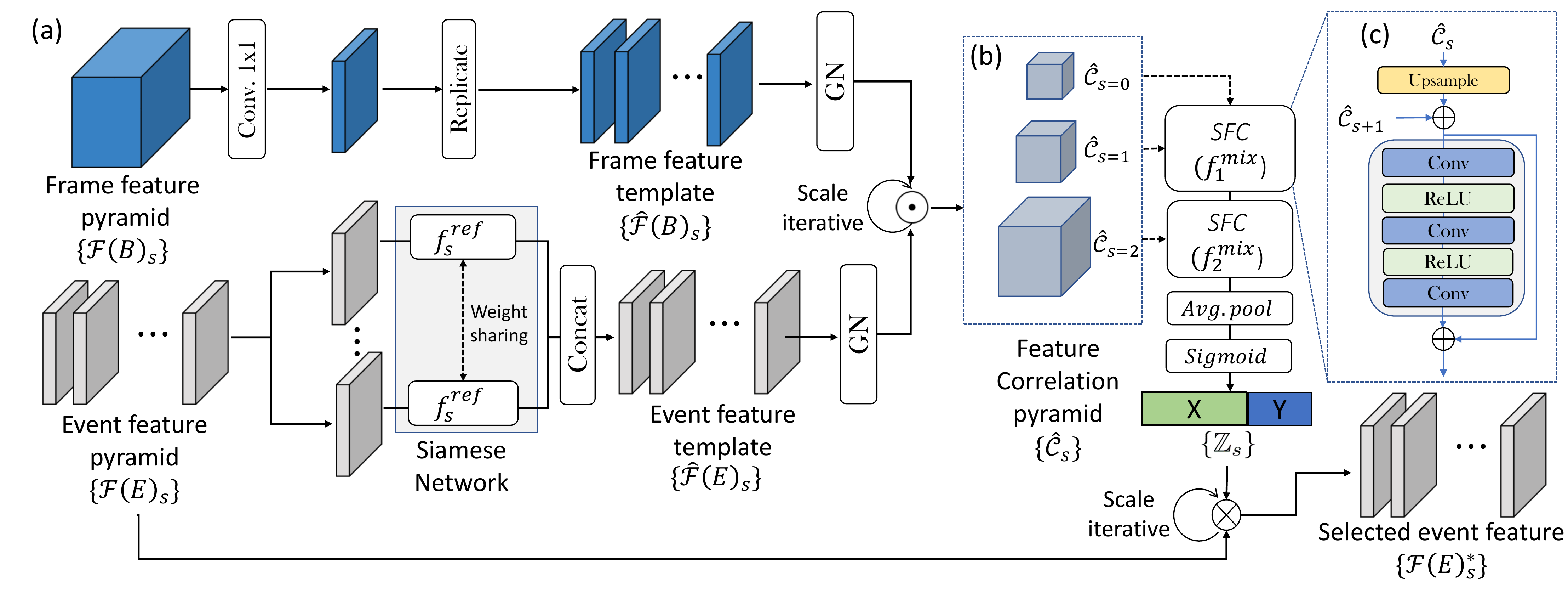} 
\caption{The network structure of the proposed Exposure-Time based Event Selection Module. ``GN'' denotes the group normalization~\cite{wu2018group}.}
\label{fig4:event_selection}
\end{figure}

\noindent \textbf{Exposure Time-based Event Selection (ETES) Module.}
Through encoding, we obtain the event and frame feature pyramids $\{\mathcal{F}(E)_{s}\}$ and $\{\mathcal{F}(B)_{s}\}$, respectively.
To approximate Eq.(\ref{eq:event_selection}), we now aim to select the beneficial event features corresponding to the dynamically varying unknown exposure time of the frame-based camera. However, there exist two crucial challenges. The first one is how to pre-process event features (with complete temporal information) and frame features (with missing temporal information).
The second is how to better discover the cross-modal relationship between events and frames by aggregating the feature pyramid obtained from two different modalities.
To this end, we propose a novel ETES module \textit{without any supervision}, as depicted in Fig.~\ref{fig4:event_selection}.
The main idea is to temporally mine the essential channels of event features based on the multi-scale cross-modal correlation.
That is, we aim to suppress the event feature corresponding to the duration of the readout phase by calculating the similarity between the blur frame feature and the event feature along with the temporal flow.
As depicted in Fig.~\ref{fig4:event_selection}(a), we first pre-process the event and frame features to calculate cross-modal correlation at multiple visual scales.
For the frame features at scale $s$, we first compress them by applying the point-wise convolution to reduce spatial information loss.
Then, we replicate the compressed frame features to have the same temporal dimension as the event features. The operations are formulated as:
\begin{equation}
    \hat{\mathcal{F}}(B)_{s} = \xi(Conv_{1\times1, s}(\mathcal{F}(B)_{s})),
\end{equation}
where $Conv_{1\times1, s}$ is the point-wise convolution at scale $s$ such that $\mathbb{R}^{C^{U}_{s} \times H_{s} \times W_{s}} \rightarrow \mathbb{R}^{C_{s=1}^{U} \times H_{s} \times W_{s}}$. Here,
$\xi$ denotes the replication operation along the temporal dimension to form frame template features $\hat{\mathcal{F}}(B)_{s}$ such that $\mathbb{R}^{C^{U}_{s=1} \times H_{s} \times W_{s}} \rightarrow \mathbb{R}^{T \times C_{s=1}^{U} \times H_{s} \times W_{s}}$.
This allows calculating the cross-modal correlation in unit time interval $\Delta t_{unit}$.
For the event features at scale $s$, we leverage a siamese network~\cite{koch2015siamese} to modulate them to effectively maintain the temporal information within $\Delta t_{unit}$, as shown in Fig.~\ref{fig4:event_selection}(a).
We apply this w.r.t. each temporal unit of the event features, $f^{\mathrm{ref}}_{s}:\mathbb{R}^{C_{s}^{U} \times H_{s} \times W_{s}} \rightarrow \mathbb{R}^{C_{s=1}^{U} \times H_{s} \times W_{s}}$.
The features are concatenated along the temporal dimension to form event template features $\hat{\mathcal{F}}(E)_s$.
We finally apply the group normalization~\cite{wu2018group} to these two template features to mitigate the extreme modality differences.
Through pre-processing both the event and frame features, we can get two feature templates at each scale $s$. As shown in Fig.~\ref{fig4:event_selection}(b), it is imperative to explore the correlations from the feature pyramids to select the most beneficial events. 
For this reason, we aggregate these two feature templates via the Hadamard product $\odot$ for all scales as
\begin{equation}
\hat{\mathcal{C}_s} = ReLU(\hat{\mathcal{F}}(E)_{s} \odot \hat{\mathcal{F}}(B)_{s})
\end{equation}
where the feature correlation $\hat{\mathcal{C}_s} \in \mathbb{R}^{T \times C_{s=1}^{U}\times H_{s}\times W_{s}}$ with ReLU for removing noisy correlation.
Finally, we get a feature correlation pyramid $\{\mathcal{C}_{s}\}$ seen from multiple visual scales.
As illustrated in Fig.~\ref{fig4:event_selection}(c), we merge the collection of correlation features by designing a scale-fusion convolutional (SFC) block.
In particular, SFC aims to form a multi-scale temporal activation map  $\{\mathbb{Z}_{s}\} \in \mathbb{R}^{T \times C_{s}^{U} \times 1 \times 1}$ in three steps. SFC upsamples the correlation features at scale $s$, followed by an element-wise addition with the features at scale $s+1$. It then propagates the most beneficial correlation information in a top-down manner. In such a way, it effectively enables the merge of lower-level to a higher-level cross-modal correlation between the frame and event.
Lastly, we squeeze the output tensor on the spatial dimension by global average-pooling followed by a sigmoid activation function. 
The output tensor is the condensed temporal activation map $\mathbb{Z}_{s=1} \in \mathbb{R}^{T \times C_{s=1}^{U} \times 1 \times 1}$.
We then interpolate $\mathbb{Z}_{s=1}$ to get the temporal activation map at each scale s $\mathbb{Z}_{s} \in \mathbb{R}^{T \times C_{s}^{U} \times 1 \times 1}$ as
\begin{equation}
\mathcal{F}(E)_{s}^{*} =  \mathbb{Z}_{s} \otimes \mathcal{F}(E)_{s}
\end{equation}
where $\otimes$ denotes channel-wise multiplication. As such, the ETES module filters and selects the event features w.r.t. unknown exposure time of the frame, as shown in Fig.~\ref{fig:time_activation_map_ETES}.
Each of the selected event feature pyramids $\{\mathcal{F}(E)^{*}_{s}\}$ is fed into the feature fusion module of each scale, as depicted in Fig.~\ref{fig2:overall_network_architecture}.

\begin{figure}[!t]
\centering
\includegraphics[width=0.7\columnwidth]{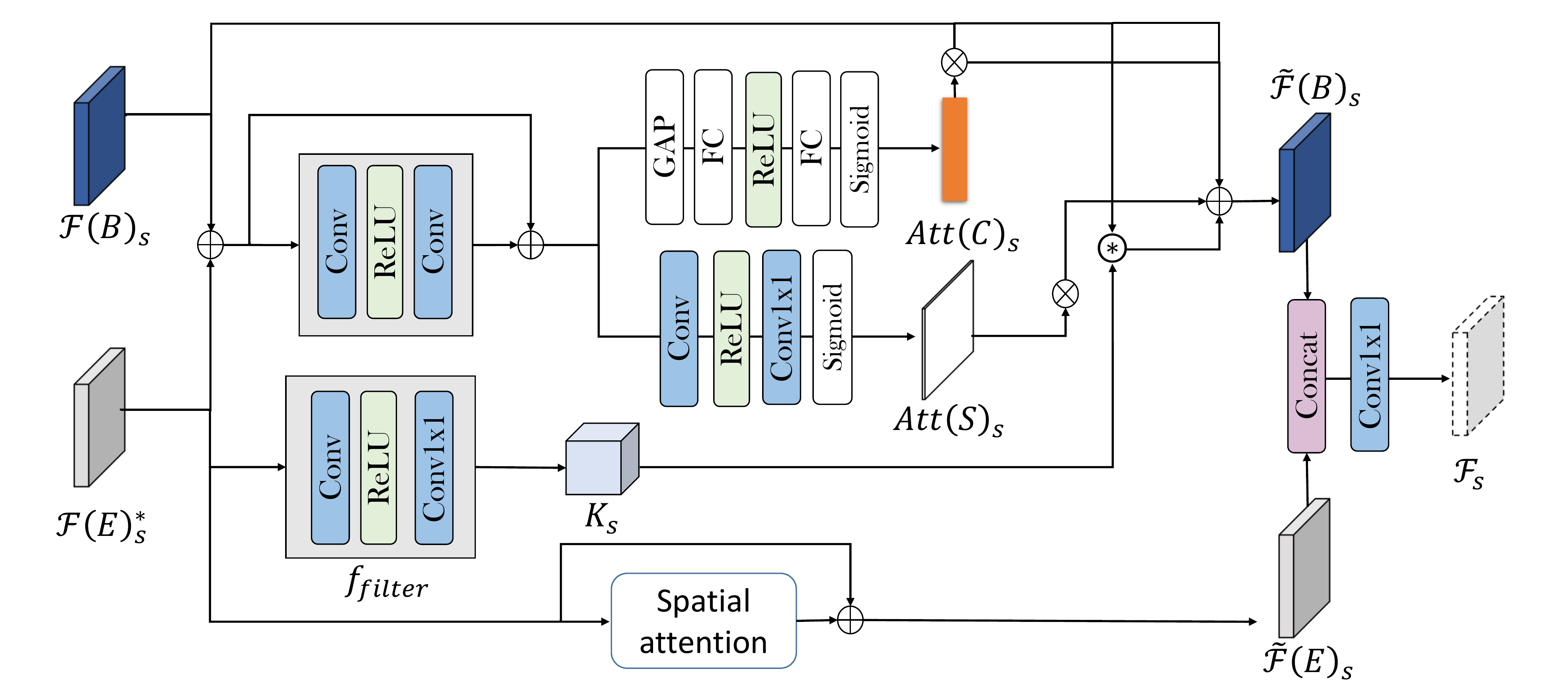} 
\caption{Proposed feature fusion module. In the figure, $\circledast$ denotes dynamic convolution operation with a generated convolution filter.}
\label{fig5:feature_fusion}
\end{figure}

\subsection{Feature Fusion Module}\label{fea_fuse_sec}
The event features $\mathcal{F}(E)_{s}^{*}$ are selected via the event selection module. We notice that the blur frame features~$\mathcal{F}(B)_{s} $ contain rich semantic and texture information, while the selected event features contain clear edge and motion cues. Therefore, it is meaningful to leverage the information from event features to complement the frame features. 
It is possible to use existing feature fusion methods directly, \eg, RGB-D~\cite{sun2021deep,zhang2020asymmetric}. However, as events are remarkably different from the frame, naively using these methods rather degrades the deblurring performance.
Consequently, we propose a novel event-frame feature fusion module for deblurring, as depicted in Fig.~\ref{fig5:feature_fusion}.
Our idea is to leverage the event features to mine the discriminative channels and spatial information from the blur frame features.
That is, we first calibrate the frame and event features via element-wise summation.
As a result, the frame features of blurry regions are highlighted through calibration due to the motion cues of the event features.
We then employ the global average pooling(GAP) to obtain global statistics, which are fed into a fully connected(FC) layer to obtain a channel attention vector for the frame features $Att(C)_{s} \in \mathbb{R}^{T \times C_{s} \times 1 \times 1}$.
Meanwhile, we extract spatial attention maps $Att(S)_{s}$ to attain the spatial statistics from calibrated features.
As such, only the important blur frame features related to deblurring are highlighted, and the unnecessary information is suppressed.
Moreover, we design a filter generation block $f_{\mathrm{filter}}$ (see Fig.~\ref{fig5:feature_fusion}) to generate a position-specific convolution filter $\mathit{K}_{s} = f_{\mathrm{filter}}(\mathcal{F}(E)^{*}_{s})$ from the selected event features $\mathcal{F}(E)^{*}_{s}$, inspired by kernel prediction networks~\cite{jia2016dynamic,niklaus2017video,mildenhall2018burst,lin2020learning}. The filtered features are formulated as:
\begin{equation}
\mathcal{\tilde{F}}(B)_{s} =  \mathcal{F}(B)_{s} + Att(C)_{s} \otimes \mathcal{F}(B)_{s} \nonumber
+ Att(S)_{s} \cdot \mathcal{F}(B)_{s} +  \mathit{K}_{s}\circledast \mathcal{F}(B)_{s}
\end{equation}
where $\circledast$ denotes convolution operation and $\otimes$ denotes channel-wise multiplication.
For event features, we only apply spatial attention~\cite{Woo_2018_ECCV} from the selected event feature
$\mathcal{F}(E)_{s}^{*}$. 
The fused features $\mathcal{F}_{s}$ at each scale are obtained by concatenating the enhanced event features $\mathcal{\tilde{F}}(E)_{s}$ and the filtered frame features $\mathcal{\tilde{F}}(B)_{s}$, followed by a 1 $\times$ 1 convolution.
Lastly, the fused feature $\{\mathcal{F}_{s}\}$ are separately fed into the decoder to reconstruct a sharp video, as shown in Fig.~\ref{fig2:overall_network_architecture}. The outputs of the decoder consist of sharp video frames at each scale, represented as $\{O_{s}\}$. We use the charbonnier loss~\cite{johnson2016perceptual} for optimization, and the total loss is:
\begin{equation}
\mathcal{L}_{total} = \sum_{s=0}^{2}\lambda_{s}\sqrt{\|O_{gt, s} - O_{s}\|^{2} + \varepsilon^{2}}
\label{loss_function}
\end{equation}
We empirically set to $\varepsilon = 10^{-3}$ for all experiments.

\section{Experiments}
\subsection{Datasets and Implementation Details}\label{sec:dataset}
\noindent \textbf{Synthetic event datasets}\; We train and test our framework on the GoPro dataset~\cite{nah2017deep}, widely used for deblurring.
We then test with the test split of the Adobe-240fps datasets~\cite{su2017deep} using the trained model on the GoPro dataset.
For both datasets, we follow an official data split, and events are generated from the high frame rate video using the event simulator (ESIM)~\cite{rebecq2018esim}. 
We synthesize blurry video by averaging the video frames.
To mimic the real video frame acquisition, we follow the method in~\cite{zhang2020video}.
We discard several video frames to simulate the readout time.
We denote the number of video frames of the exposure phase as $m$ and that of the readout phase as $n$.
We downsample the original video from 240fps to 15fps with $m+n = 16$.
We set the frame number $m$ of the exposure phase as an odd number $m$ =\{9, 11, 13, 15\}.
During training, we add random noise, $\epsilon \sim [-0.6n, 0.6n]$, to the readout interval for better generalization.
Accordingly, we can simulate the random video frame acquisition process in training.
\begin{table*}[t]
\centering
\caption{Quantitative evaluation on a synthetic event dataset. Asterisk(*) means retraining our training dataset. \textit{$\dagger$ denotes the event-guided method.} The \textbf{Bold} and \uline{underline} denote the best and the second-best performance, respectively. We trained our method on the GoPro dataset and directly applied it to the Adobe dataset. The same notation and typography are applied to the following tables.}\label{tab:synthetic_quantitative}
\scalebox{0.67}{
\begin{tabular}{ccccccccccccccccc}
\thickhline
\multicolumn{1}{c}{\multirow{3}{*}{Method}} & \multicolumn{8}{c}{GoPro -15fps}                                                                                        & \multicolumn{8}{c}{Adobe-15fps}                                                                                   \\ \cline{2-17} 
\multicolumn{1}{l}{}                        & \multicolumn{2}{c}{GoPro-9-7} & \multicolumn{2}{c}{GoPro-11-5} & \multicolumn{2}{c}{GoPro-13-3} & \multicolumn{2}{c|}{GoPro-15-1}         & \multicolumn{2}{c}{Adobe-9-7} & \multicolumn{2}{c}{Adobe-11-5} & \multicolumn{2}{c}{Adobe-13-3} & \multicolumn{2}{c}{Adobe-15-1} \\ \cline{2-17} 
\multicolumn{1}{l}{}                        & PSNR       & SSIM       & PSNR        & SSIM       & PSNR        & SSIM       & PSNR & \multicolumn{1}{c|}{SSIM} & PSNR         & SSIM        & PSNR        & SSIM        & PSNR        & SSIM       & PSNR        & SSIM       \\ \hline
Nah \etal ~\cite{nah2017deep} & 28.89 & 0.930 & 27.76 & 0.914 & 26.70 & 0.895 & 25.74 & 0.876 &
28.29 & 0.914 & 27.35 & 0.900 & 26.59 & 0.888 & 25.96 & 0.878 \\
DMPHN~\cite{zhang2019deep} & 31.21 & 0.946 & 30.39 & 0.936 & 28.75 & 0.914 & 26.83 & 0.880 &
29.17 & 0.920 & 28.22 & 0.905 & 27.21 & 0.888 & 26.31 & 0.872 \\
Kupyn \etal ~\cite{Kupyn_2019_ICCV} & 30.18 & 0.932 & 29.10 & 0.918 & 28.11 & 0.903 & 27.21 & 0.888 &
29.81 & 0.924 & 28.85 & 0.913 & 28.09 & 0.903 & 27.51 & 0.894 \\
DBGAN~\cite{zhang2020deblurring} & 32.15 & 0.955 & 31.44 & 0.945 & 29.25 & 0.921 & 26.99 & 0.881 & 29.91
& 0.927 & 28.82 & 0.912 & 27.70 & 0.895 & 26.69 & 0.878 \\
BANet~\cite{tsai2021banet} & 33.02 & 0.961 & 32.38 & 0.956 & 30.89 & 0.941 & 28.93 & 0.915 & 31.01
& 0.943 & 29.96 & 0.929 & 28.84 & 0.913 & 27.90 & 0.898 \\ 
MPRNet~\cite{zamir2021multi} & 33.77 & 0.967 & 32.65 & 0.959 & 31.24 & 0.946 & 29.66 & 0.927 & 31.20 & 0.945 & 30.17 & 0.933 & 29.05 & 0.919 & 28.16 & 0.906 \\ 
MIMO-UNet+~\cite{cho2021rethinking} & 33.34 & 0.964 & 32.39 & 0.956 & 30.74 & 0.940 & 28.52 & 0.908 &
30.83 & 0.939 & 29.76 & 0.925 & 28.57 & 0.908 & 27.51 & 0.892 \\
HINet~\cite{chen2021hinet} & 33.60 & 0.965 & 32.86 & 0.960 & 31.26 & 0.945 & 29.09 & 0.917 & 31.06
& 0.944  & 30.08 & 0.932 & 28.90 & 0.915 & 27.91 & 0.901 \\ 
ESTRNN$^{*}$ ~\cite{zhong2020efficient} & 29.97 & 0.929 & 28.93 & 0.916 & 27.96 & 0.901 & 27.04 & 0.885 & 28.36 & 0.907 & 27.45 & 0.893 & 26.72 & 0.881 & 26.10 & 0.870 \\
CDVD-TSP~\cite{pan2020cascaded} & 29.13 & 0.926 & 28.53 & 0.917 & 27.77 & 0.905 & 26.89 & 0.890 & 27.57 & 0.907  & 27.21 & 0.900 & 26.80 & 0.892 & 26.43 & 0.885 \\  \hline
Nah \etal$^{\dagger *}$~\cite{nah2017deep} & 34.39 & 0.966 & 34.37  & 0.965 & 34.04 & 0.963 & 33.66 & 0.961 & 33.63  & 0.962 & 33.64  & 0.962  & 33.10  & 0.958  & 32.25  & 0.950  \\ 
DMPHN$^{\dagger *}$~\cite{zhang2019deep} & 33.68 &  0.961 & 33.63 & 0.961 & 33.36 & 0.959 & 33.03 & 0.956 & 33.16 & 0.959 & 33.02  & 0.959 & 32.55 & 0.955 & 32.01  & 0.949  \\ 
LEDVDI$^{\dagger *}$~\cite{lin2020learning} & 33.39 & 0.958 & 33.57 & 0.959 & 33.68 & 0.959 & 32.81 & 0.953  & 32.92 & 0.958 & 32.92 & 0.958 & 33.18  & 0.953 & 31.91 & 0.949 \\
D2Nets$^{\dagger *}$~\cite{Shang_2021_ICCV} & 29.52 & 0.923 & 29.39 & 0.921 & 29.56 & 0.921 & 29.04  & 0.911 & 28.41 & 0.909 & 28.24 & 0.905 & 28.46 & 0.906 & 28.14 & 0.900 \\ \hline
\textbf{Ours-light$^{\dagger}$} & \uline{34.91} & \uline{0.969} & \uline{34.61} & \uline{0.968} & \uline{34.43} & \uline{0.966} & \uline{34.00} & \uline{0.964} & \uline{34.19} & \uline{0.965} & \uline{33.72} & \uline{0.963} & \uline{33.87} & \uline{0.964} & \uline{33.20} & \uline{0.959}
\\
\textbf{Ours$^{\dagger}$} & \textbf{36.22} & \textbf{0.976} & \textbf{35.93} & \textbf{0.974} & \textbf{35.67} & \textbf{0.973} &  \textbf{35.28} & \textbf{0.971} & \textbf{35.52} & \textbf{0.973} & \textbf{35.24} & \textbf{0.971} & \textbf{35.11} & \textbf{0.970}  & \textbf{34.67} & \textbf{0.968} \\ \thickhline
\end{tabular}}
\end{table*}
As such, we get a synthetic dataset that simulates various exposure times.
We denoted this dataset as ``dataset-m-n''. \\
\noindent \textbf{Real-world event datasets}\;
To evaluate our method on real-world events, we collected 53,601 sharp images of 59 different scenes with slow-motion using the DAVIS 346 color event camera that provides aligned events and RGB data($346\times260$ resolution).
We attain the blurry video by averaging the sharp frames with the setting $m+n=10$ for network training.
We set the frame number of the exposure phase as an odd number $m$ =\{3, 5, 7, 9\}.
Similarly, we add a random noise $\epsilon \sim [-0.6n, 0.6n]$ to the read-out interval in the training phase.
Finally, we generated training sets consisting of blur images and corresponding sharp ground truths images with events for 43 scenes.
For testing, \emph{\textbf{we set the test set configuration differently from the composition of the training sets}} by setting $m+n=14$ and $m=\{9,11,13\}$ to confirm the generalization ability for the unseen video frame acquisition process. 
Finally, we generated a test set consisting of 3,588 blur and GT images with events for 16 scenes.
In this manner, we conduct a quantitative evaluation with other methods.
In addition, we collected real-world blurry videos by shooting various scenes with fast motion to evaluate our method on the real-world blurry videos qualitatively.
For the real-world blurry video shooting, we set the exposure time as $\{15, 25, 35, 45, 55\}$ms or auto-exposure and the shutter period as 60ms.
Then, we conducted experiments on real-world blurry videos at various unknown exposure times.
\\
\noindent\textbf{Implementation Details} \;
Our frameworks are implemented using PyTorch~\cite{paszke2017automatic}.
For all datasets, we utilize the batch size of 8 and ADAM~\cite{kingma2014adam} optimizer to update weight using a multi-step scheduler with an initial learning rate $1e^{-4}$ and decay rate $\gamma=0.5$.
$\lambda_{s}$ of Eq.(\ref{loss_function}) are set to $\{1, 0.1, 0.1\}$ for each scale. 
For data augmentation, we apply random cropping($256 \times 256$) to the event and frame for the same position. We adopt the dynamic convolution operation from the STFAN~\cite{Zhou_2019_ICCV} implementation. 
For quantitative evaluation, we use common evaluation metrics PSNR and SSIM~\cite{wang2004image}.

\begin{table}[t]
\caption{Quantitative evaluation and complexity analysis on real event dataset. The inference time and FLOPs are measured using TITAN RTX GPU on 346 × 260 resolution images of test sets. $\ddagger$ means event-guided deblurring methods evaluated using the official pretrained model. As their official models only provide the result of a grayscale image input, we evaluate the performance of each model on the grayscale image input.}~\label{tab:real_world_quantitative}
\centering
\scalebox{0.75}{
\begin{tabular}{c|cccccccc|c}
\thickhline
\multirow{3}{*}{Methods} & \multicolumn{8}{c|}{Real-world event dataset} & \multirow{3}{*}{\begin{tabular}[c]{@{}c@{}}Complexity\\ FLOPs(G)/Runtime(ms)\end{tabular}} \\ \cline{2-9}
 & \multicolumn{2}{c}{9-5} & \multicolumn{2}{c}{11-3} & \multicolumn{2}{c|}{13-1} & \multicolumn{2}{c|}{Avg.} &  \\ \cline{2-9}
 & PSNR & SSIM & PSNR & SSIM & \multicolumn{1}{l}{PSNR} & \multicolumn{1}{l|}{SSIM} & PSNR & SSIM &  \\ \hline
Nah \etal$^{*}$~\cite{nah2017deep} & 29.74 & 0.8420 & 28.74 & 0.8207 &  27.96 & \multicolumn{1}{c|}{0.8037} & 28.81 & 0.8221 & 245.40/154.65  \\
DMPHN$^{*}$~\cite{zhang2019deep} & 29.76 & 0.8392 & 28.80 & 0.8188 &  28.03 & \multicolumn{1}{c|}{0.8021} & 28.87 & 0.8200 & 56.99/27.99  \\
CDVD-TSP$^{*}$~\cite{pan2020cascaded} & 32.95 & 0.9077 & 31.73 & 0.8878 & 30.66 &  \multicolumn{1}{c|}{0.8683} & 31.78 & 0.8880 & 281.15/204.46 \\ 
MPRNet$^{*}$~\cite{zamir2021multi} & 30.42 & 0.8596 & 29.22 & 0.8345 & 28.25 & \multicolumn{1}{c|}{0.8136} & 29.30 & 0.8359 & 1247.17/151.9  \\
MIMO-UNet+$^{*}$~\cite{cho2021rethinking} & 30.32 & 0.8529 & 29.24 & 0.8309 & 28.37 & \multicolumn{1}{c|}{0.8129} & 29.31 & 0.8322 &  112.44/44.0 \\ \hline
e-SLNet$^{\ddagger}$~\cite{wang2020event} & 20.82 & 0.6379 & 21.39 & 0.6603 & 22.09 & \multicolumn{1}{c|}{0.6872} & 21.43 & 0.6546 & 114.88/55.87  \\
REDS$^{\ddagger}$~\cite{Xu_2021_ICCV} & 26.12 & 0.7399 & 30.20 & 0.8448 & 31.34 & \multicolumn{1}{c|}{0.8542} & 29.22 & 0.8130 & 116.89/47.15 \\ \hline
Nah \etal$^{\dagger*}$~\cite{nah2017deep} & 35.10 & 0.9326 & 32.84 & 0.9057 & 32.41 &  \multicolumn{1}{c|}{0.9016} & 33.45  & 0.9133 & 247.00/156.32  \\
DMPHN$^{\dagger*}$~\cite{zhang2019deep} & 33.87 & 0.9069 & 33.09 & 0.8958 & 33.02 & \multicolumn{1}{c|}{0.8957} & 33.33 & 0.8995 & 57.88/27.22  \\
LEDVDI$^{\dagger*}$~\cite{lin2020learning} & 34.77 & 0.9258 & 33.83 & 0.9138 & 32.96 &  \multicolumn{1}{c|}{0.9047} & 33.86 & 0.9148 &  62.80/25.24 \\ 
D2Nets$^{\dagger*}$~\cite{Shang_2021_ICCV} & 31.36 & 0.8753 & 30.87 & 0.8663 & 29.90 &  \multicolumn{1}{c|}{0.8481} & 30.71 & 0.8632 &  283.86/243.9 \\ \hline
\multicolumn{1}{c|}{Ours-light$^{\dagger}$} & \multicolumn{1}{c}{\uline{35.53}} & \multicolumn{1}{c}{\uline{0.9342}} & \multicolumn{1}{c}{\uline{34.58}} & \multicolumn{1}{c}{\uline{0.9232}} & \multicolumn{1}{c}{\uline{34.64}} & \multicolumn{1}{c|}{\uline{0.9248}} & \multicolumn{1}{c}{\uline{34.92}} & \multicolumn{1}{c|}{\uline{0.9274}} & \multicolumn{1}{c}{60.72/32.14} \\
\multicolumn{1}{c|}{Ours$^{\dagger}$} & \multicolumn{1}{c}{\textbf{36.98}} & \multicolumn{1}{c}{\textbf{0.9487}} & \multicolumn{1}{c}{\textbf{36.10}} & \multicolumn{1}{c}{\textbf{0.9407}} & \multicolumn{1}{c}{\textbf{35.98}} & \multicolumn{1}{c|}{\textbf{0.9404}} & \multicolumn{1}{c}{\textbf{36.35}} & \multicolumn{1}{c|}{\textbf{0.9433}} & \multicolumn{1}{c}{237.77/75.07} \\ 
\thickhline
\end{tabular}}
\end{table}

\subsection{Experimental Results}
\noindent 
We compare with Nah \etal~\cite{nah2017deep} and DMPHN~\cite{zhang2019deep} by feeding the RGB frame with embedded events to the networks (denoted as Nah \etal$^{\dagger}$, DMPHN$^{\dagger}$).
For comparison of the SOTA event-guided video deblurring method D2Nets$^{\dagger}$~\cite{Shang_2021_ICCV}, we used the official training code.
In addition, we reimplement the other SOTA event-guided video deblurring methods~\cite{lin2020learning}(denoted as LEDVDI$^{\dagger}$) based on the code provided by the authors.
We keep the original network architecture and with modification of  the event representation~\cite{zhu2019unsupervised}.

\noindent\textbf{Synthetic event datasets} For a fair comparison, we \textit{retrain} the one frame-based video deblurring method~\cite{zhong2020efficient} and four event-guided methods(D2Nets$^{\dagger}$, LEDVDI$^{\dagger}$ and Nah \etal$^{\dagger}$ and DMPHN$^{\dagger}$) on our training dataset. Also, we compare with the SOTA frame-based methods using official pre-trained models
 provided by the authors ~\cite{pan2020cascaded,nah2017deep,zhang2019deep,zhang2020deblurring,Kupyn_2019_ICCV,tsai2021banet,zamir2021multi,cho2021rethinking,chen2021hinet}.
As clearly shown in Tab.~\ref{tab:synthetic_quantitative}, our method surpasses the frame-based and event-guided methods by a large margin on the two datasets.
Compared to the frame-based method, the avg. PSNR score of our method improves from 3.95dB to 8.50dB in the GoPro-15fps, from 5.49dB to 8.09dB in the Adobe-15fps.
As we maximize the number of video frames corresponding to the exposure phase, the performance gap between our method and frame-based competitors widens from 5.62dB to 9.54dB in the GoPro-15-1.
This indicates our method achieves better results on challenging motion blur frames.
Compared to the event-guided method, our method still shows better results from 1.66 dB to 6.40dB in the GoPro-15fps and 1.98 to 6.82dB in the Adobe-15fps.

\begin{figure}[t]
\centering
\includegraphics[width=.95\columnwidth]{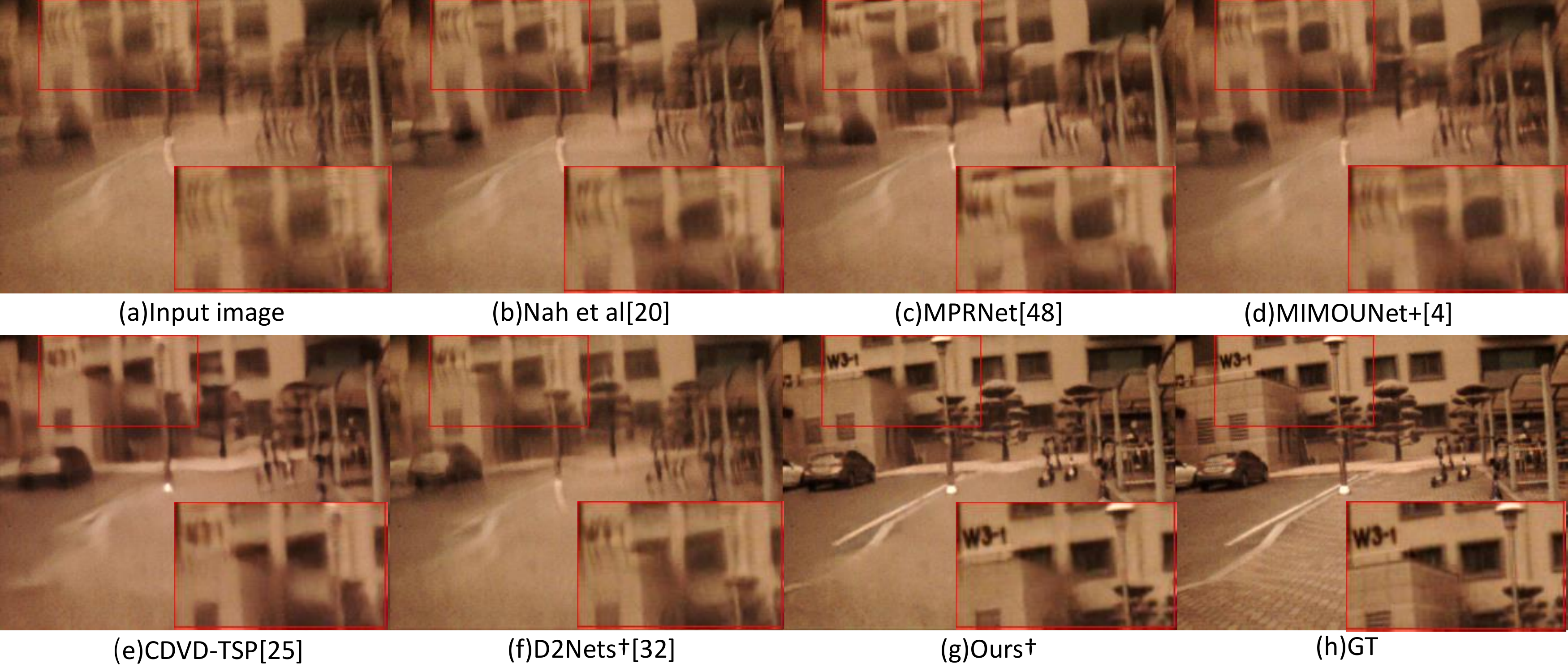}
\caption{Visual comparison on the test split of real-world event datasets.}\label{fig:qualitative_real_synthetic_dvs}
\end{figure}
\begin{figure}[!t]
\centering
\includegraphics[width=0.8\columnwidth]{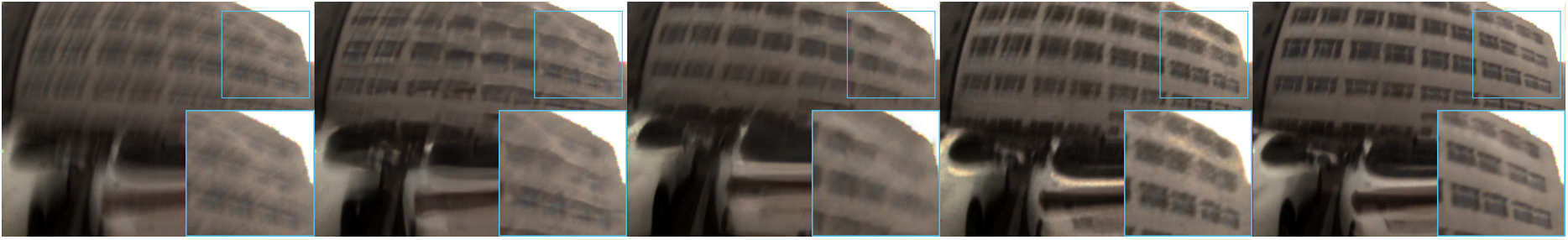}
\includegraphics[width=0.8\columnwidth]{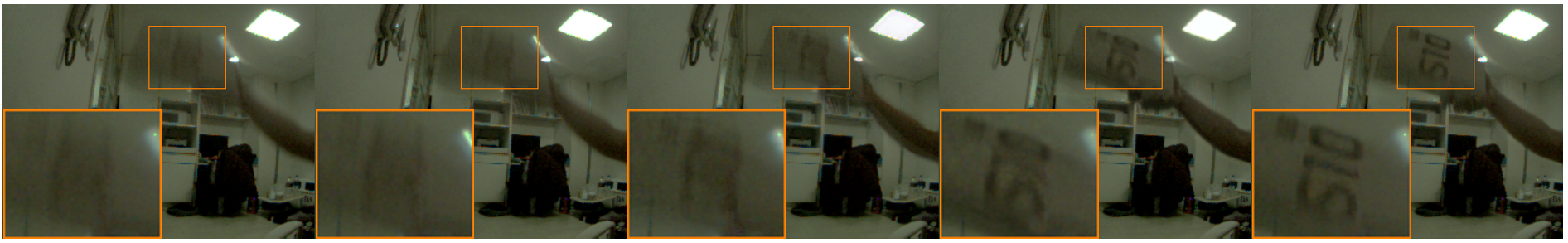}
\caption{Deblurring results on \textbf{real-world} unknown exposure time blurry video frames. From left to right: inputs, MIMO-UNet+~\cite{cho2021rethinking}, D2Nets$^{\dagger}$~\cite{Shang_2021_ICCV}, LEDVDI$^{\dagger}$~\cite{lin2020learning}, Ours$^{\dagger}$. }\label{fig:qualititave_real_blur}
\end{figure}
\begin{figure}[t]
  \centering
  \includegraphics[width=0.75\columnwidth]{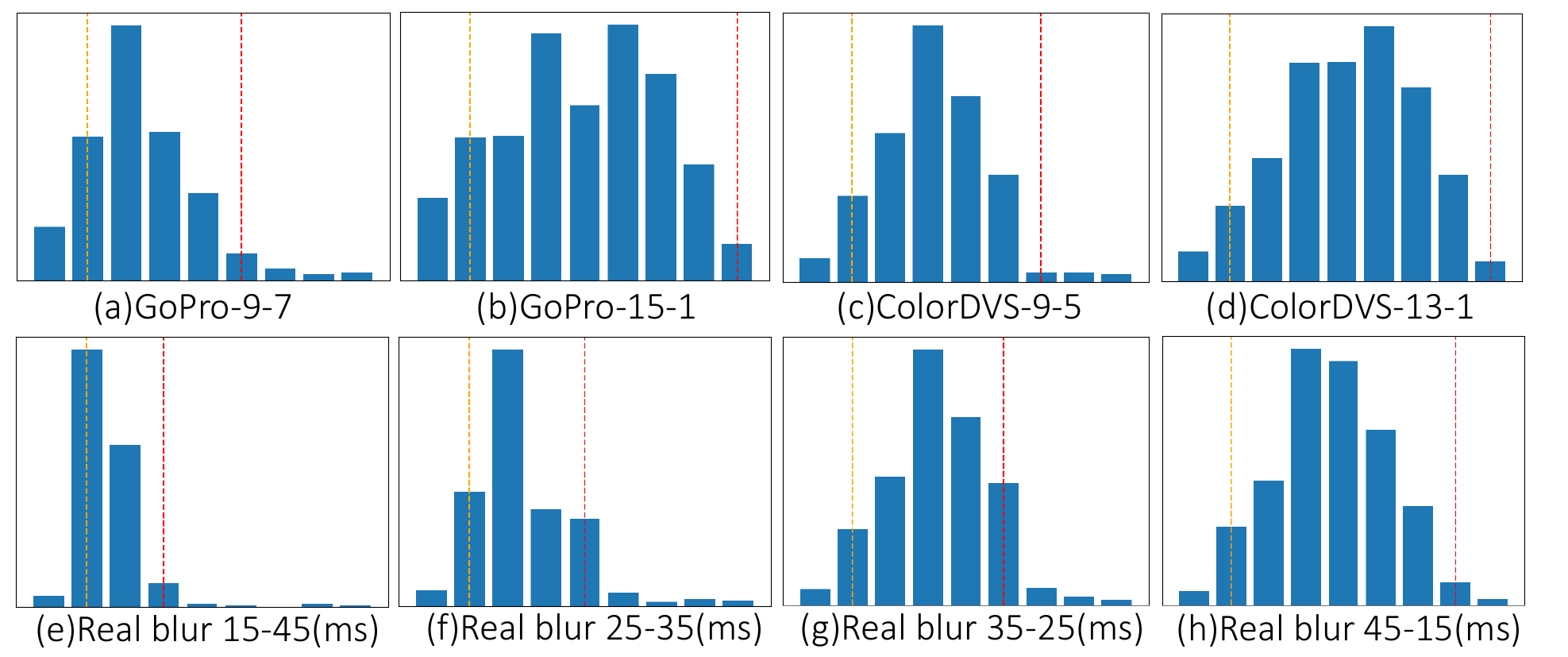}
  \caption{The visualization of the average temporal activation map of the ETES module on various datasets, including real-shot blurry videos. The horizontal and vertical axes represent the temporal axis and the average amount of channel activation, respectively. The yellow and red dotted lines indicate the start and end of the exposure time, respectively. In the second row,
 the first and the last numbers indicate the exposure and readout time of real-world blurry videos.
 }\label{fig:time_activation_map_ETES}
\end{figure}

\begin{table}[t]
\centering
\caption{The ablation study of the individual components.} \label{tab:ablation_study}
\scalebox{0.95}{
\begin{tabular}{c|c|c|c|c|c|c|c|c}
\thickhline
RE &  &  & \checkmark & \checkmark & \checkmark & \checkmark & \checkmark  & \checkmark  \\
MS &  & \checkmark &  & \checkmark &  &  & \checkmark  & \checkmark \\
ETES &  &  &  &  & \checkmark & \checkmark &  \checkmark & \checkmark \\
FF &  &  &  &  &  & \checkmark &   & \checkmark  \\ \hline
PSNR & 34.45 & 34.93 & 34.96 & 35.42 & 35.77 & 36.09 & \underline{36.02} & \textbf{36.35} 
\\ \thickhline
\end{tabular}}
\end{table}

\noindent\textbf{Real-world event datasets}\; 
As mentioned above, the real-world datasets contain various scenes different from the synthetic event dataset, so we retrain the SOTA frame-based methods~\cite{pan2020cascaded,zhang2019deep,nah2017deep,zamir2021multi,cho2021rethinking} and event-guided methods(LEDVDI$^{\dagger}$, Nah \etal$^{\dagger}$, DMPHN$^{\dagger}$, D2Nets$^{\dagger}$) on our real-world event datasets.
In addition, we evaluate the SOTA event-guided deblurring methods(e-SLNet$^{\ddagger}$~\cite{wang2020event},
REDS$^{\ddagger}$~\cite{Xu_2021_ICCV}) using official pretrained model provided by the authors.
In Tab.~\ref{tab:real_world_quantitative}, we reported the deblurring performance of unseen and unknown exposure time videos compared with other methods.
Compared to the frame-based methods, there is a performance gap(\textbf{4.57} to \textbf{7.54} dB) on avg. PSNR and SSIM(\textbf{0.0553} to \textbf{0.1233}). 
Moreover, our network records a significant performance improvement compared to the event-guided methods(retrained in our datasets) on avg. PSNR(\textbf{2.49} to \textbf{5.64} dB) and SSIM(\textbf{0.0285} to \textbf{0.0801}) with comparable test cost and running time.
The “ours-light” network, the light version of our original network with reduced channels and res-blocks, has higher performance (1.06dB to 4.21dB) with low computational cost. Ours-light network shows the slight difference of test costs and running time with two light event-guided methods (DMPHN$^{\dagger}$~\cite{zhang2019deep}, LEDVDI$^{\dagger}$~\cite{lin2020learning}) and faster and lighter than the other two event-guided methods (D2Nets$^{\dagger}$~\cite{Shang_2021_ICCV}, Nah \etal$^{\dagger}$~\cite{nah2017deep}) with distinct performance differences.
In Fig.~\ref{fig:qualitative_real_synthetic_dvs}, only our method effectively restores a detailed structure even under extreme non-linear motion.
\\
\noindent\textbf{Evaluation results on real-world blurry video frames}\;
Finally, we experiment generalization ability of our methods on real-world blurry videos. For testing, real-shot blurry video frames and corresponding events (in 1/(shutter speed) duration) are given to our model inputs. Then, we checked the results for the temporal activation map of our ETES module according to various unknown exposure times to confirm the inference ability of exposure time for deblurring. Specifically, we plot the average distribution of the activated event channels for each temporal unit of the ETES module by averaging over 200 real-world blurry frames for each video clip in Fig.~\ref{fig:time_activation_map_ETES}(e), (f), (g), (h). \textit{\textbf{We confirm all temporal activations are concentrated near the exposure phase even in the real-blurry video frames without exact exposure time information.}} This indicates that event features are well selected based on cross-modality feature correlations according to real-world blurry videos’ exposure time. Lastly, we performed qualitative comparisons with other methods(D2Nets†~\cite{Shang_2021_ICCV} , LEDVDI†~\cite{lin2020learning}, MIMO-UNet+~\cite{cho2021rethinking}) as illustrated in Fig.~\ref{fig:qualititave_real_blur}. 
Only our method can restore the letters written on the box in the second row.
\subsection{Ablation study}
\noindent We analyzed the performance contribution of our network modules.
All ablation experiments are performed on the real-world event dataset with models trained for $2.5\times 10^5$ iterations.

\noindent\textbf{Recurrent Encoding (RE)} 
To demonstrate the effectiveness of RE, we compare the model using 2D CNNs and the proposed shared-RNN cell for embedding event streams.
From the 1st and 3nd rows of Tab.~\ref{tab:ablation_study}, we observe a performance gain (+0.51db) when using RE.
On the other hand, when using the 2D CNNs for encoding events, temporal information lying in the event streams is not well preserved, adversely affecting deblurring.
Furthermore, if we use a multi-scale loss function(denoted as ``MS''), we observe a performance gain (+0.46db).

\noindent\textbf{ETES module}
is the most crucial module in our method.
We validate the deblurring performance with and without this module to verify the effectiveness.
From the 3rd and 5th columns of Tab.~\ref{tab:ablation_study}, 
we observe a significant performance improvement (+0.81db) in terms of the average PSNR.
We plot the average distribution of the ETES module's output according to various unknown exposure times using all video frames of the test sets in the synthetic event datasets(Fig.~\ref{fig:time_activation_map_ETES}(a), (b)) and real-world event datasets(Fig.~\ref{fig:time_activation_map_ETES}(c), (d)).
Here, we confirm that all activations are within the exposure phase and hardly activate at the readout phase, even in the different configuration in the training phase. 
This shows the \textit{efficacy} of the ETES module.

\noindent\textbf{Feature Fusion Module}\;
For cross-modality feature fusion, the simplest way is to use a concatenation of two features to fuse two different modality features. 
In Tab.~\ref{tab:ablation_study}, by comparing 7rd and 8th columns, we can observe performance improvement (+0.33db). 
The proposed feature fusion module can better utilize the complementary information from frame and event features.
\section{Conclusion}
\noindent  This paper studied and formulated a new research problem of event-guided motion deblurring for unknown exposure time videos. To this end, we proposed a novel end-to-end framework. Specifically, we proposed a method of selectively using event features by estimating the feature correlation of different modalities of events and frames. Moreover, extensive experiments demonstrated our method significantly surpasses existing event-guided and frame-based deblurring methods on the various datasets, including real-world blurry videos. \\
\textbf{Acknowledgements} This work was supported by Institute of Information and Communications Technology Planning \& Evaluation(IITP) Grant funded by Korea Government (MSIT) (No. 2020-0-00440, Development of Artificial Intelligence Technology that Continuously Improves Itself as the Situation Changes in the Real World and No.2014-3-00123, Development of High Performance Visual BigData Discovery Platform for Large-Scale Realtime Data Analysis) and the National Research Foundation of Korea(NRF) grant funded by the Korea government(MSIT) (NRF2022R1A2B5B03002636)

\clearpage
%
%
\bibliographystyle{splncs04}
\bibliography{egbib}

\newpage
\section{Appendix}
\subsection{Additional related work}
\noindent\textbf{Deep learning for Event-to-Image and Video Reconstruction.}
The other line of research directly reconstructs sharp images and video from event data via adversarial learning~\cite{Wang_2019_CVPR,mostafavi2021learning,zhang2020learning}, RNNs~\cite{rebecq2019events,zou2021learning,stoffregen2020reducing}, and self-supervised learning~\cite{paredes2020back}. As the event cameras, \eg, DAVIS 240C~\cite{brandli2014240}, are in a low-resolution, some attempts have tried to reconstruct high-resolution images via supervised \cite{mostafavi2020learning} and unsupervised learning~\cite{Wang_2020_CVPR}. Moreover, some works~\cite{wang2021evdistill,wang2021dual} have demonstrated that image reconstruction can be used to help event-based visual perception tasks in training.
However, reconstructing video from the events is still a highly ill-posed problem due to inherently unstable contrast threshold and sensor noise.

\noindent\textbf{Cross-Modal Attention}\;
Attention mechanisms can adaptively transform a network’s parameters according to inputs. Thus, it boosts representative features while suppressing uninformative features in various manners, such as channel attention, spatial attention, and temporal attention~\cite{Hu_2018_CVPR,Wang_2017_CVPR,Chen_2017_CVPR,park2018bam,Woo_2018_ECCV,Li_2019_CVPR}.
Recently, a growing body of research has been delving into dynamic feature modulation considering two modality inputs. 
For event and frame modalities, Gehrig \etal~\cite{gehrig2021combining} introduce a recurrent feature modulation mechanism to fuse the event and RGB sensor data.

\begin{figure}[!b]
\centering
\includegraphics[width=0.5\columnwidth]{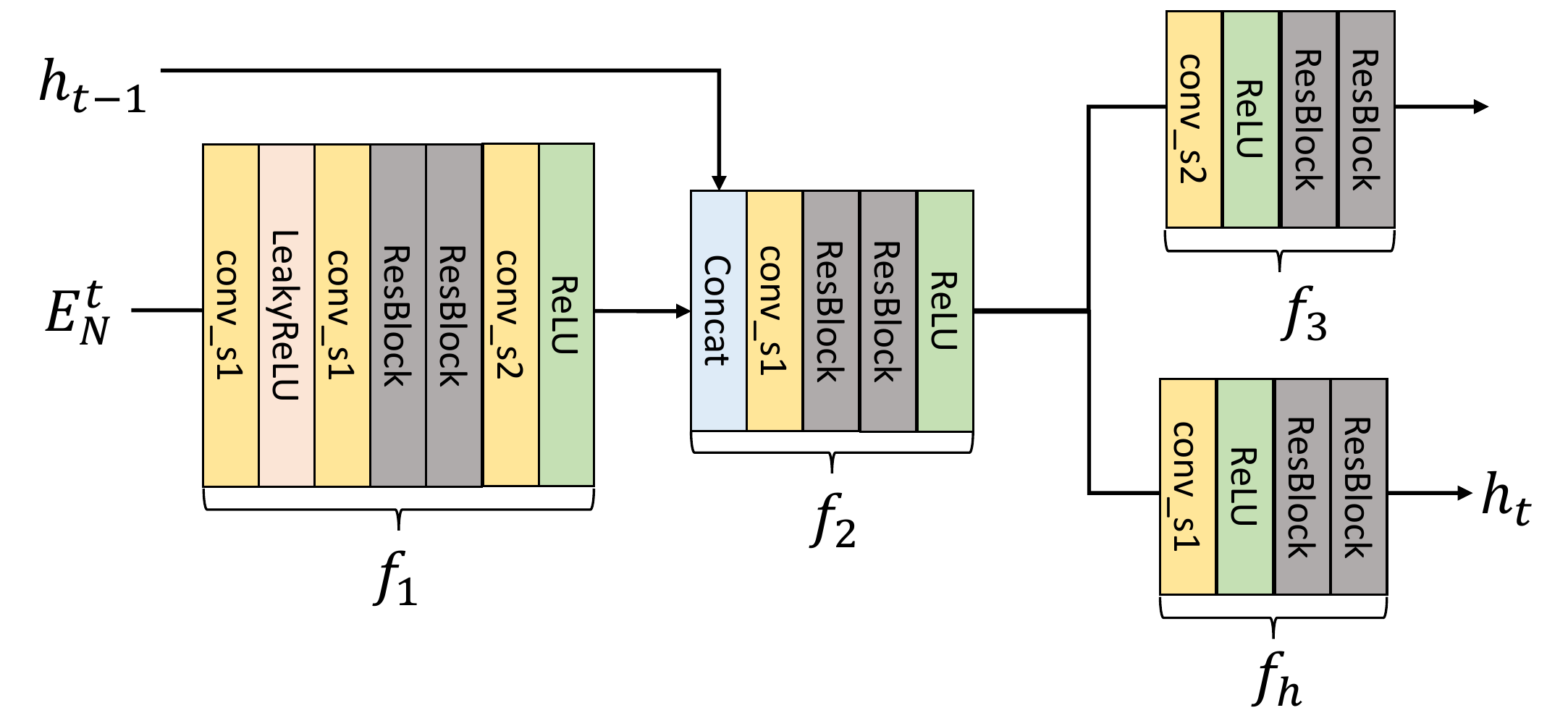}
\caption{The proposed RNN-cell for recurrent encoding of the events}
\label{fig:RNN_cell_ECCV}
\end{figure}

\subsection{Network Architecture}
\subsubsection{Recurrent encoding for the embedded events}
To apply the proposed RNN-cell, we first divided $B$ temporal bins of the voxel grid of the events into $N$ temporal units as mentioned in the main paper.
For each temporal-wise divided unit event $\{E_{t}^n \} \in \mathbb{R}^{2 \times H \times W}$ with temporal index $n\in\{1, ..., N\}$, we apply the proposed RNN cell as illustrated Fig.\ref{fig:RNN_cell_ECCV}.
We first extract the feature of the unit event utilizing the first encoding block $f_{1}$ as follows:
\begin{equation}
    \mathcal{F}(E_{t}^{n})_{s=0} = f_{1}(E_{t}^{n})
\end{equation}
We then generate a feature map of the next scale.
\begin{equation}
    \mathcal{F}(E_{t}^{n})_{s=1} = f_{2}(Concat(\mathcal{F}(E_{t}^{n})_{s=0}, h_{t-1}))
\end{equation}
where $Concat$ denotes channel-wise concatenation operation; $f_{2}$ refers to the second CNN block of the RNN cell, and $h_{t-1}$ refers to the previously generated hidden state.
With these local event feature $\mathcal{F}(E_{t}^{n})_{s=1}$, we recursively update hidden state $h_{t}$ as follows:
\begin{equation}
    h_{t} = f_{h}(\mathcal{F}(E^{n}_{t})_{s=1})
\end{equation}
where $f_{h}$ denotes the CNNs block for extracting the hidden state. We then further process $\mathcal{F}(E^{n}_{t})_{s=1}$ using the last CNNs block $f_{3}$ represented as:
\begin{equation}
    \mathcal{F}(E_{t}^{n})_{s=2} = f_{3}(\mathcal{F}(E^{n}_{t})_{s=1})
\end{equation}
In this way, we generate the output hierarchical feature maps $\{\mathcal{F}(E_{t}^{1})_{s}, ..., \mathcal{F}(E_{t}^{N})_{s}\}$ for the current part($s\in\{0,1,2\}$).
All the generated feature maps are concatenated with the feature map of events for the past part.

\begin{table}[!t]
\centering
\caption{Dataset comparison between Blur-DVS and our datasets}\label{tab:dataset_comparison} 
\scalebox{1.0}{
\begin{tabular}{|c|c|c|c|c|c|}
\hline
 & Camera & Resolution & Color & \# sharp images & \# Scenes \\ \hline
Blur-DVS~\cite{jiang2020learning,lin2020learning,Shang_2021_ICCV} & DAVIS-240C & $240 \times 180 $ & No & 15246 & -  \\ 
Our datasets & Color DAVIS-346 & $ 346 \times 260$  & Yes  & 53601 & 59 \\ \hline
\end{tabular}
}
\end{table}

\begin{figure}[!t]
\centering
\includegraphics[width=0.3\columnwidth]{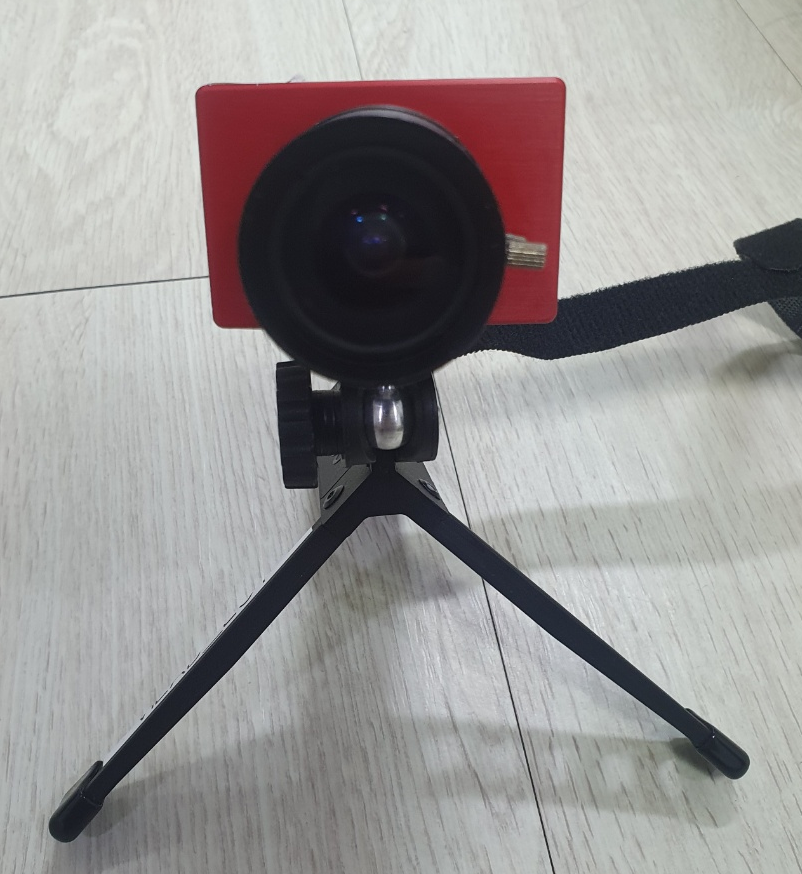}
\caption{DAVIS-346 Color camera used for dataset collection.}
\label{fig:DAVIS_COLOR}
\end{figure}
\subsection{Real-world event datasets}
\subsubsection{Dataset collection details and comparisons}
As mentioned in the main paper, there are no publicly available large-scale datasets for evaluating event-guided motion deblurring, including real-world events.
The previous event-guided motion deblurring methods used the dataset called named as Blur-DVS~\cite{jiang2020learning,lin2020learning,Shang_2021_ICCV}, which is not publicly available.
For this reason, we collected a new dataset using the Color-DAVIS 346 camera that provides RGB images and spatially aligned stream of events as shown in Fig.\ref{fig:DAVIS_COLOR}.
Since the Color-DAVIS 346 camera has a low frame rate (maximum $\sim \mathrm{40fps}$), we captured static scenes when collecting sharp images.
We minimize motion blur by moving the camera slowly.
Compared to Blur-DVS~\cite{jiang2020learning,lin2020learning,Shang_2021_ICCV}, we collected more sharp images and diverse scenes, as shown in Table.\ref{tab:dataset_comparison}.
In addition, we obtained relatively higher resolution events and images pair since the Color-DAVIS 346 camera can shoot events and frames with a higher resolution than the DAVIS-240C.
Finally, our dataset contains RGB images, whereas Blur-DVS~\cite{jiang2020learning,lin2020learning,Shang_2021_ICCV} only provides intensity images.
Therefore, we can evaluate richer texture information of the scene details using RGB images.

\begin{table}[!t]
\centering
\caption{The evaluation results of our model on the GoPro dataset when tested in different configurations from the training set using three different training datasets. We trained our network for the number of iterations the same as used in the main paper.}\label{tab:dataset_generation}
\scalebox{0.85}{
\begin{tabular}{|c|cccccccc|}
\hline
\multirow{3}{*}{Training datasets} & \multicolumn{8}{c|}{Unseen interval} \\ \cline{2-9} 
 & \multicolumn{2}{c}{7-5} & \multicolumn{2}{c}{9-3} & \multicolumn{2}{c|}{11-1} & \multicolumn{2}{c|}{Avg.} \\ \cline{2-9} 
 & PSNR & SSIM & PSNR & SSIM & PSNR & \multicolumn{1}{c|}{SSIM} & PSNR & SSIM \\ \hline
Known exposure time & 23.26 & 0.7354 & 25.03 & 0.7881 & 31.26  & \multicolumn{1}{c|}{0.9263} & 26.52 & 0.8166 \\
Unknown exposure time & 37.39  & 0.9700 & 34.62 & 0.9541 & 34.94 &  \multicolumn{1}{c|}{0.9584} &  35.65 & 0.9608 \\
Unknown exposure time + random noise & 37.07 & 0.9686 & 36.84 & 0.9669 & 36.41  & \multicolumn{1}{c|}{0.9642} & 36.77 & 0.9666  \\ \hline
\end{tabular}
}
\end{table}
\subsubsection{The experiments on dataset generations}
For performing event-guided motion deblurring for unknown exposure time videos, it is crucial to simulate various unknown video frame acquisition processes during the training phase(arbitrary exposure and readout time).
The reason is that there can be arbitrary different ratios for exposure and readout interval in real situations. 
Therefore, our ETES module needs to learn select event features corresponding to unknown exposure time at various arbitrary interval ratios.
To this end, we add random noise to readout-interval for the generalization ability.
To demonstrate the effectiveness of our proposed dataset generation method, we trained our model with three different training datasets and tested our model not seen in the training set.
First, we train our model with the assumption of the previous event-guided motion deblurring methods(shutter period and exposure time are the same - known exposure assumption). 
In that case, the performance is dramatically degraded to unseen combination of exposure and read-out time, as shown in the first row of Table.\ref{tab:dataset_generation}.
Next, we generated a training sets with $m+n=16$, set the number of video frames of the exposure phase $m$=\{9,11,13,15\} as in the main paper.
We then train our network without adding random noise to evaluate the performance.
We observe the deblurring performance is somewhat improved in the unseen interval, but the still degraded performance(the 2nd row of Table.\ref{tab:dataset_generation}).
Lastly, we observe a significant performance improvement to the unseen interval by adding random noise as shown in the 3rd row of Table.\ref{tab:dataset_generation}. 
This experiment demonstrated that we could improve the generalization ability in the unseen exposure-readout intervals by adding random noise to the readout interval.

\subsection{Additional experimental results and details}
\subsubsection{More visual comparisons on real-world unknown exposure time blurry video frames}
In Fig.~\ref{fig:real_blur_1} and Fig.~\ref{fig:real_blur_2} and Fig.~\ref{fig:real_blur_3}, we perform qualitative comparisons with the SoTA frame-based image deblurring methods
(MIMOUNet+~\cite{cho2021rethinking}) and the SoTA event-guided video deblurring methods(D2Nets~\cite{Shang_2021_ICCV}) on real-world blurry video frames captured by Color DAVIS-346 event camera.
We confirm that our method restores more precise, sharp details than other methods, even in real-world blurry video frames.
\subsubsection{More visual comparisons on the test split of our real-world event datasets}
In Fig.~\ref{fig:color_dvs_1} and Fig.~\ref{fig:color_dvs_2} and Fig.~\ref{fig:color_dvs_3} and Fig.~\ref{fig:color_dvs_4}, we perform qualitative comparisons with the SoTA frame-based image deblurring methods(DMPHN~\cite{zhang2019deep}, MIMOUNet+~\cite{cho2021rethinking}, MPRNet~\cite{zamir2021multi}) and the SoTA video deblurring method(CDVD-TSP~\cite{pan2020cascaded}) and
the SoTA event-guided video deblurring method
(D2Nets~\cite{Shang_2021_ICCV}).
We confirm that our method can more precisely restore sharp images even in severe blurry conditions caused by non-linear motion.
\subsubsection{More visual comparisons on GoPro-15fps dataset~\cite{nah2017deep}}
In Fig.~\ref{fig:GOPRO_1} and Fig.~\ref{fig:GORPO_2} and Fig.~\ref{fig:GORPO_3} and Fig.~\ref{fig:GORPO_4}, we perform qualitative comparisons with the SoTA frame-based image deblurring methods(HINet~\cite{chen2021hinet}, MIMOUNet+~\cite{cho2021rethinking}, MPRNet~\cite{zamir2021multi}) and video deblurring methods(ESTRNN~\cite{zhong2020efficient}, CDVD-TSP~\cite{pan2020cascaded}).
Our proposed networks can restore a more plausible and sharp image than other methods.

\begin{figure}[!t]
\centering
\includegraphics[width=0.8\columnwidth]{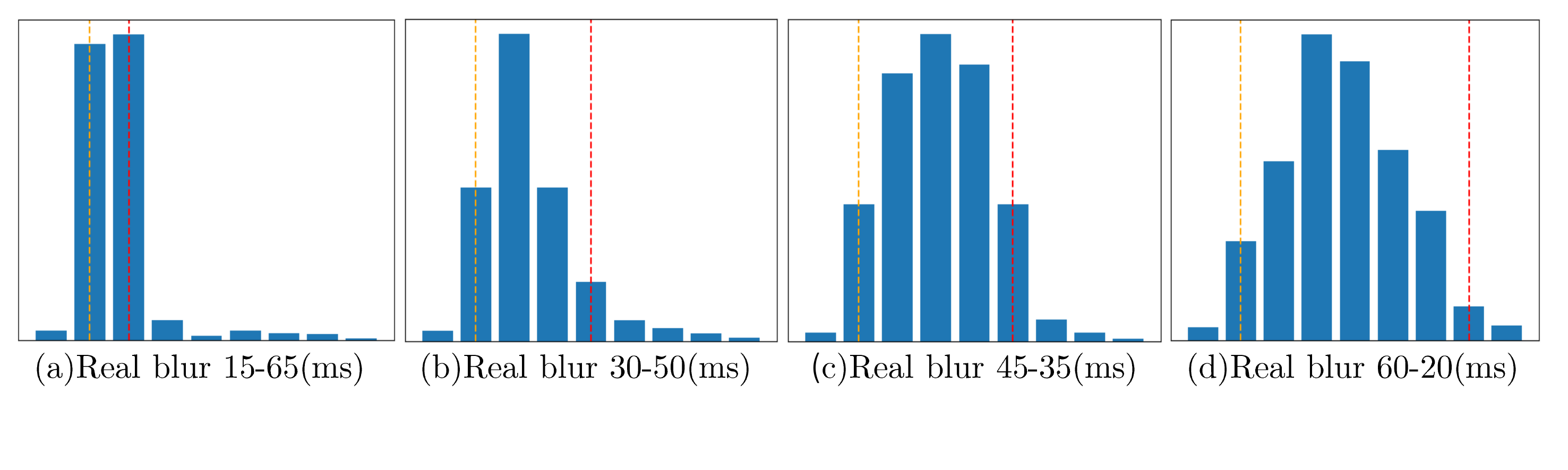}
\caption{The visualization results of the average temporal activation map of the ETES module on the real-world blurry videos. The horizontal and vertical axes
represent the temporal axis and the average amount of channel activation, respectively. The yellow and red dotted lines indicate the
start and end of the exposure time, respectively.
The first and the last numbers indicate the exposure and readout time of real-world blurry videos.}
\label{fig:ETES_ECCV}
\end{figure}
\begin{table}[!b]
\centering
\caption{The evaluation results of LEDVDI$^{\dagger}$~\cite{lin2020learning} on the real-world event dataset using the pretrained model and retrained model. Please note the performance of retrained model is much better than when using pretrained model.}\label{tab:LEDVDI}
\scalebox{0.9}{
\begin{tabular}{|c|cccccc|cc|}
\hline
\multirow{2}{*}{} & \multicolumn{2}{c}{9-5} & \multicolumn{2}{c}{11-3} & \multicolumn{2}{c|}{13-1} & \multicolumn{2}{c|}{Avg.} \\ \cline{2-9} 
 & PSNR & SSIM & PSNR & SSIM & PSNR & SSIM & PSNR & SSIM \\ \hline
LEDVDI$^{\dagger}$~\cite{lin2020learning}(pretrained) & 23.54 & 0.6783 & 23.44 & 0.6751 & 24.21 & 0.6986 & 23.73 & 0.6840 \\
LEDVDI$^{\dagger}$~\cite{lin2020learning}(retrained) & 34.77 & 0.9258 & 33.83 & 0.9138 & 32.96 & 0.9047 & 33.86 & 0.9148 \\ \hline
\end{tabular}}
\end{table}

\subsubsection{Additional average temporal activation maps of our ETES modules on real-world blurry videos}
In the main paper, we experimented with the ETES module's estimation results of unknown exposure time on real-world blurry videos.
In addition, we experimented on the exposure time estimation result of our ETES module for motion deblurring with a different combination of the exposure time-readout time setting included in the main paper.
For this experiment, we set exposure time as \{15, 30, 45, 60\}ms and shutter period as 80ms.
We then plotted averaged temporal activation map of the ETES module for 200 video frames of each video clip in Fig.\ref{fig:ETES_ECCV}.
Here, we confirmed that all activation is hardly activated in the readout phase and mainly in the exposure phase, even in the different compositions of main paper.

\subsubsection{Implementation details of other event-guided methods}
\noindent We retrain other event-guided methods(LEDVDI$^{\dagger}$~\cite{lin2020learning}, DMPHN$^{\dagger}$~\cite{zhang2019deep}, Nah \etal$^{\dagger}$~\cite{nah2017deep}, D2Nets$^{\dagger}$~\cite{Shang_2021_ICCV}) for same iterations as with our method.
For all datasets, we utilize the batch size of 8 and ADAM~\cite{kingma2014adam} optimizer to update weight using a multi-step scheduler with an initial learning rate of $1\times 10^{-4}$ and decay rate of $0.5$.
For data augmentation, we apply random cropping($256\times 256$) to the event and frame for the same
position.
Since D2Nets$^{\dagger}$~\cite{Shang_2021_ICCV} uses network input for ground truth sharp frame(non-consecutively blurry frames assumption), we replace ground truth sharp frame with blurry frame for fair comparisons.
As can be seen in the case of LEDVDI$^{\dagger}$~\cite{lin2020learning}(Tab.\ref{tab:LEDVDI}),
the retrained model shows much better results than the official pretrained model.
\subsubsection{Implementation details of other frame-based methods}
As with the event-guided methods, we retrain frame-based image deblurring method(MPRNet~\cite{zamir2021multi}, MiMOUNet+~\cite{cho2021rethinking}, DMPHN~\cite{zhang2019deep}, Nah \etal~\cite{nah2017deep}) and 
video deblurring method
(CDVD-TSP~\cite{pan2020cascaded}) for $3.75\times 10^5$ iterations in the real-world event dataset under their original hyperparameter setting provided by authors.
We apply random cropping($256 \times 256)$ to the frame during training.
For all frame-based deblurring methods training, we used the official GitHub code provided by authors.

\newpage

\begin{figure}[!t]
\centering
\includegraphics[width=0.85\columnwidth]{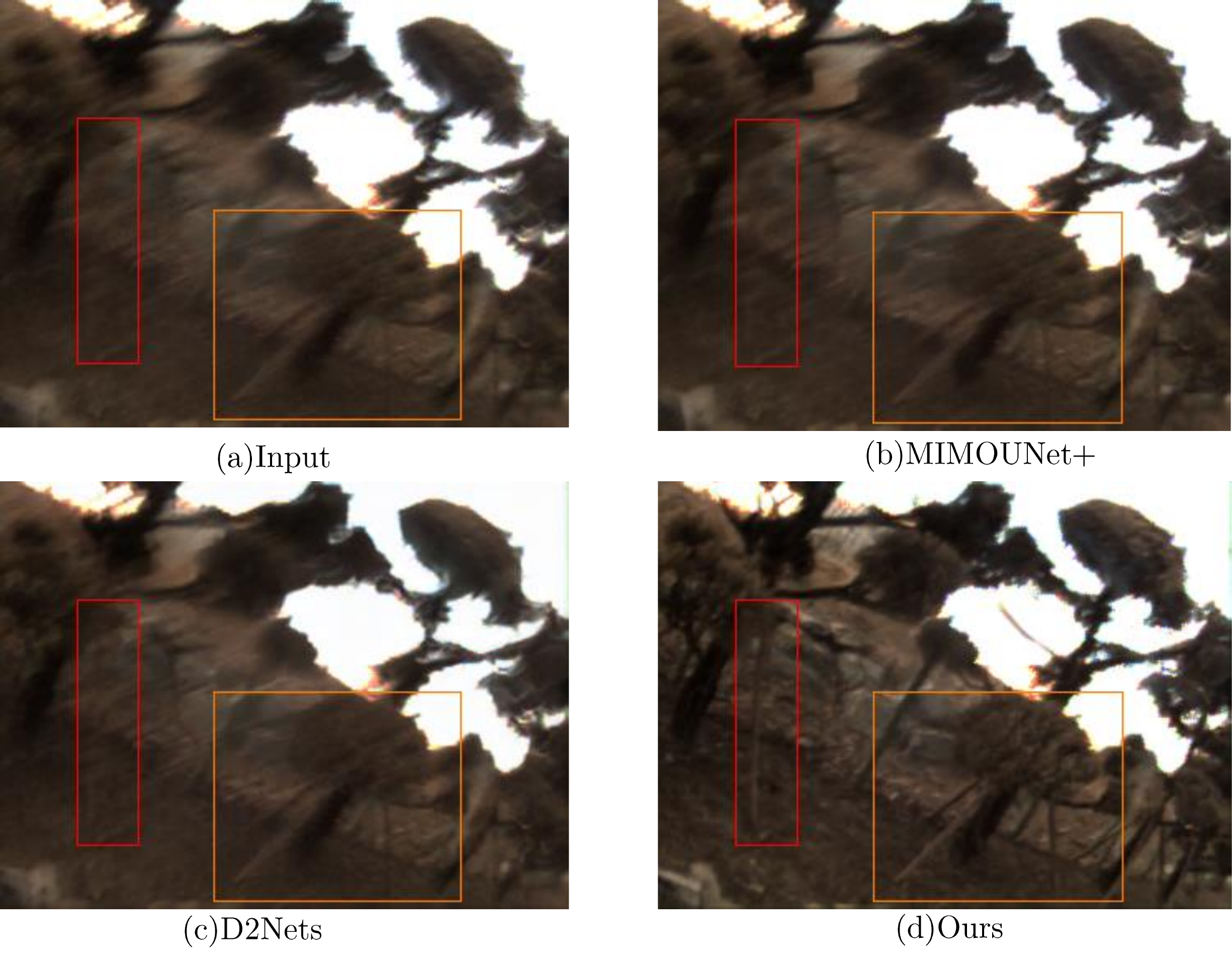}
\caption{Visual comparisons on \textbf{real-world} unknown exposure time blurry video frames.}
\label{fig:real_blur_1}
\end{figure}

\begin{figure}[!t]
\centering
\includegraphics[width=0.85\columnwidth]{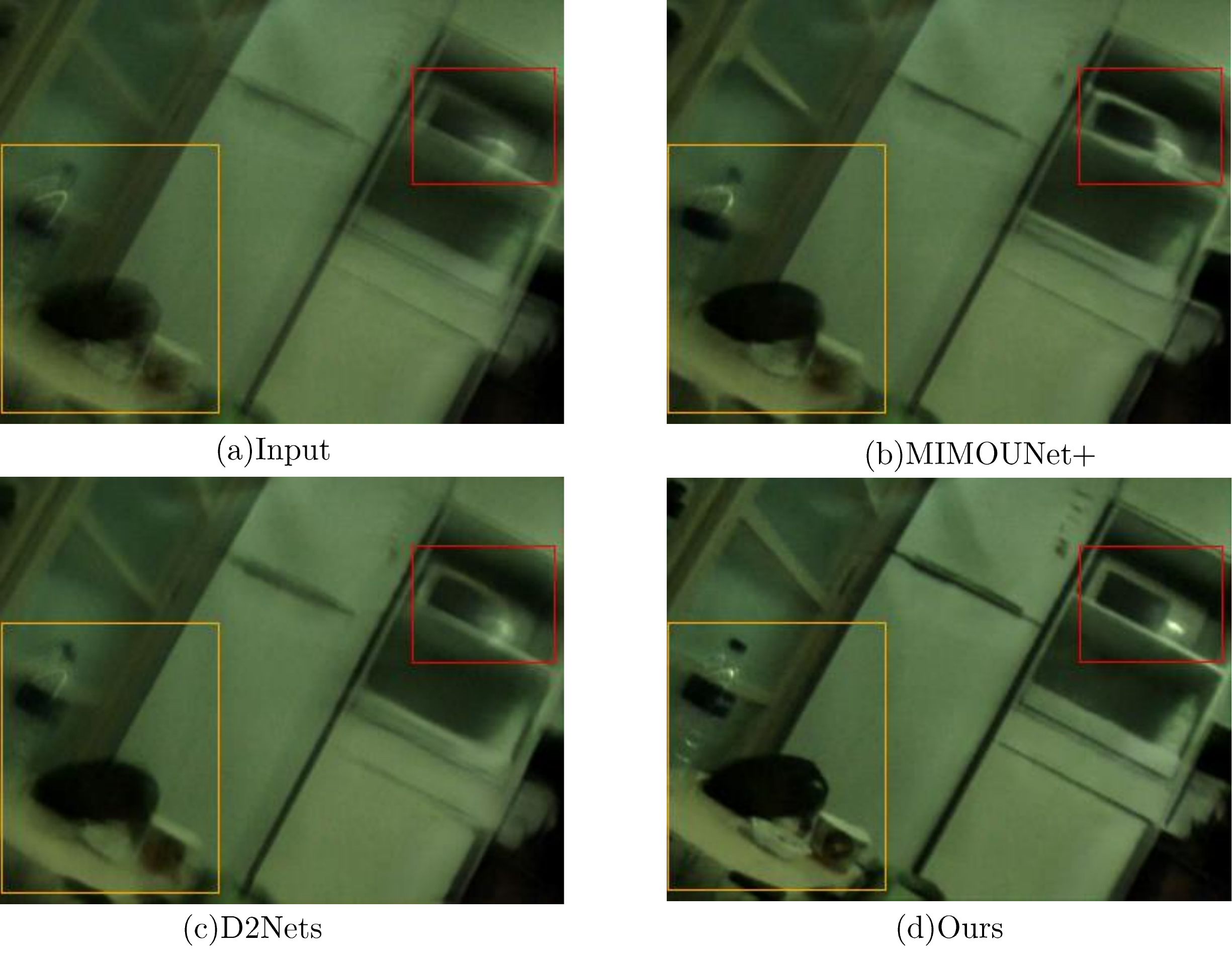}
\caption{Deblurring results on \textbf{real-world} unknown exposure time blurry video frames.}
\label{fig:real_blur_2}
\end{figure}

\newpage
\begin{figure}[!t]
\centering
\includegraphics[width=0.85\columnwidth]{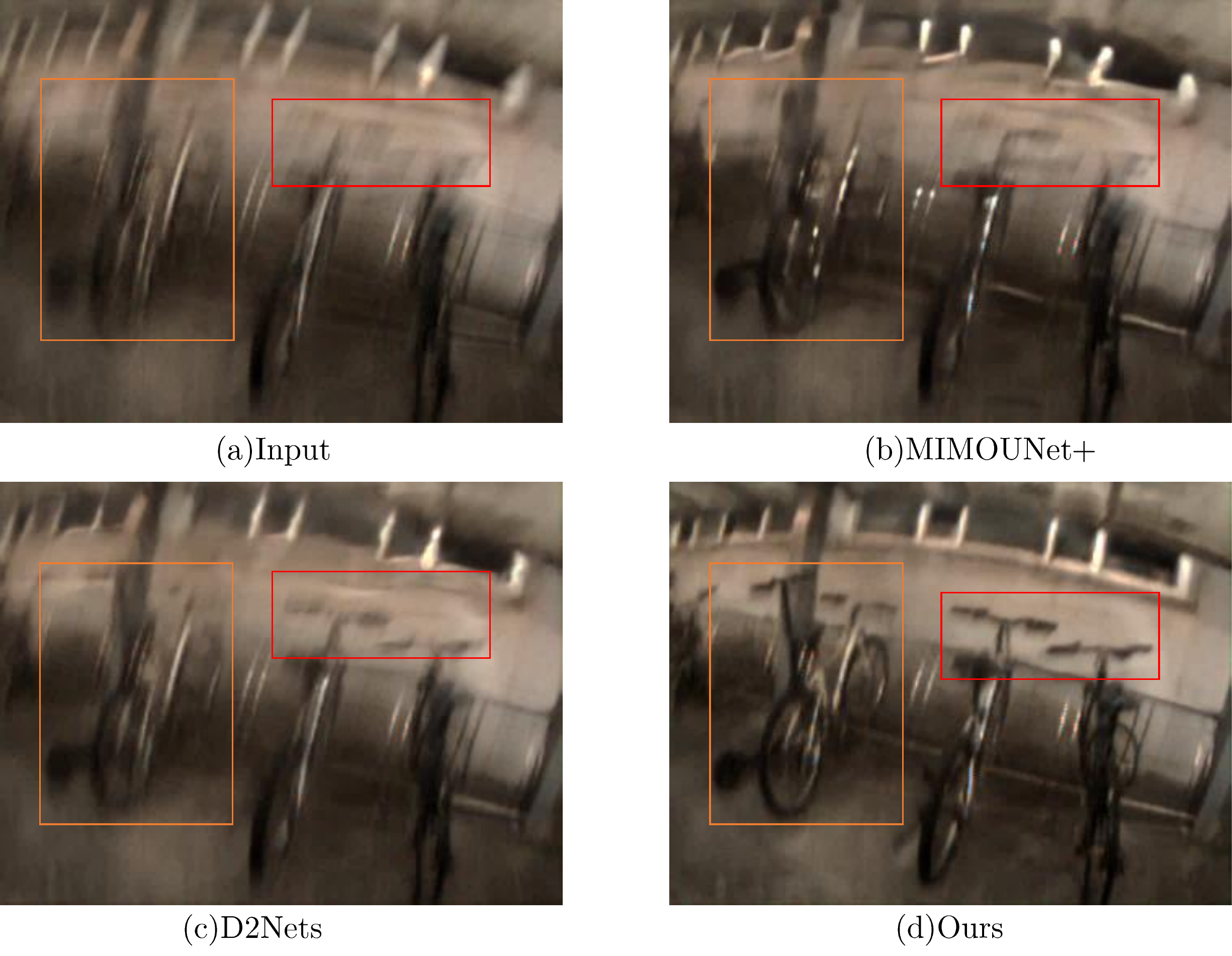}
\caption{Deblurring results on \textbf{real-world} unknown exposure time blurry video frames.}
\label{fig:real_blur_3}
\end{figure}

\newpage
\begin{figure}[!b]
\centering
\includegraphics[width=0.9\columnwidth]{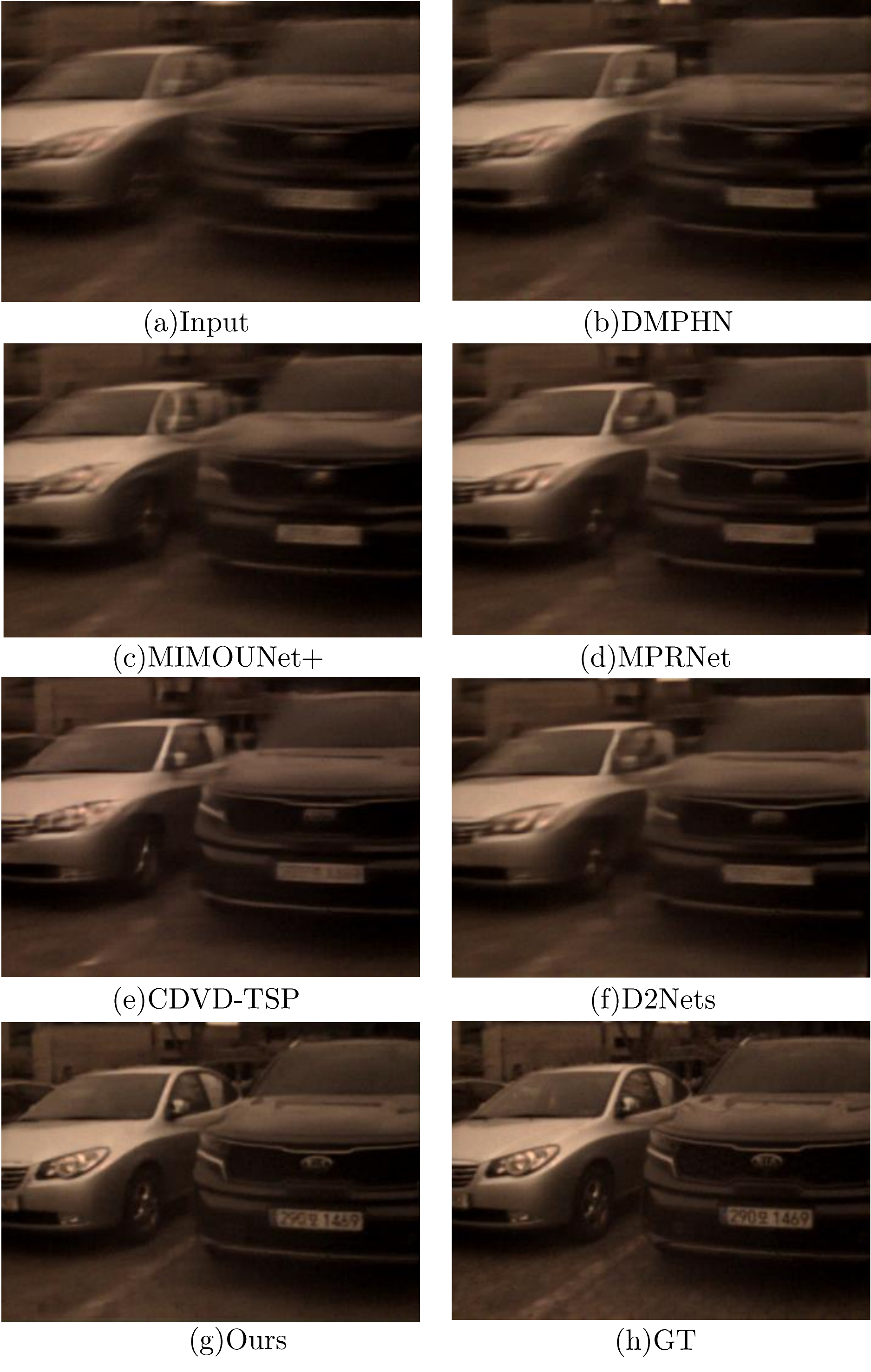}
\caption{Visual comparison of unknown exposure time blurry video frames on our real-world event datasets.}
\label{fig:color_dvs_1}
\end{figure}

\newpage
\begin{figure}[!b]
\centering
\includegraphics[width=0.9\columnwidth]{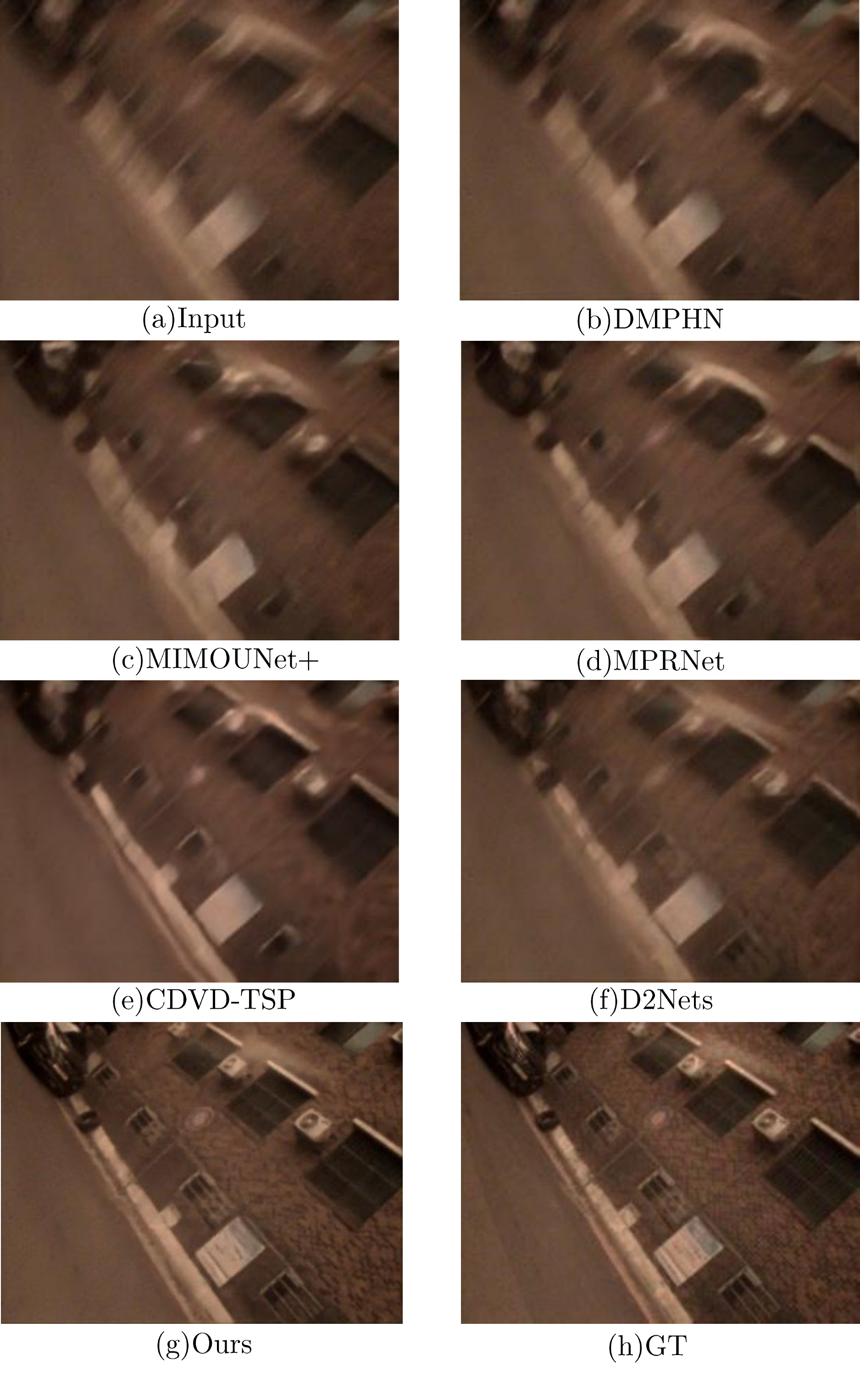}
\caption{Visual comparison of unknown exposure time blurry video frames on our real-world event datasets.}
\label{fig:color_dvs_2}
\end{figure}

\newpage

\begin{figure}[!b]
\centering
\includegraphics[width=0.9\columnwidth]{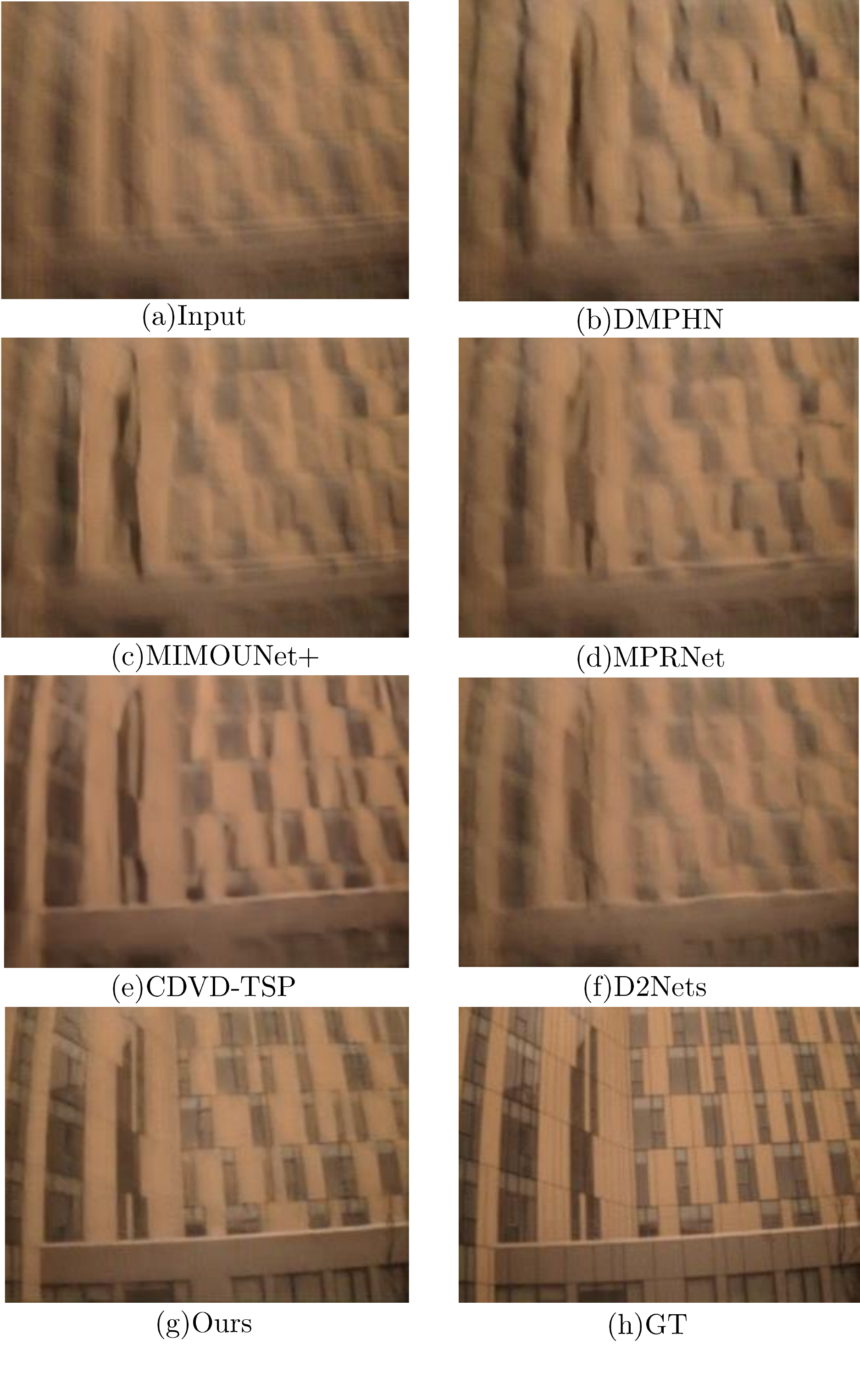}
\caption{Visual comparison of unknown exposure time blurry video frames on our real-world event datasets.}
\label{fig:color_dvs_3}
\end{figure}

\newpage
\begin{figure}[!b]
\centering
\includegraphics[width=0.9\columnwidth]{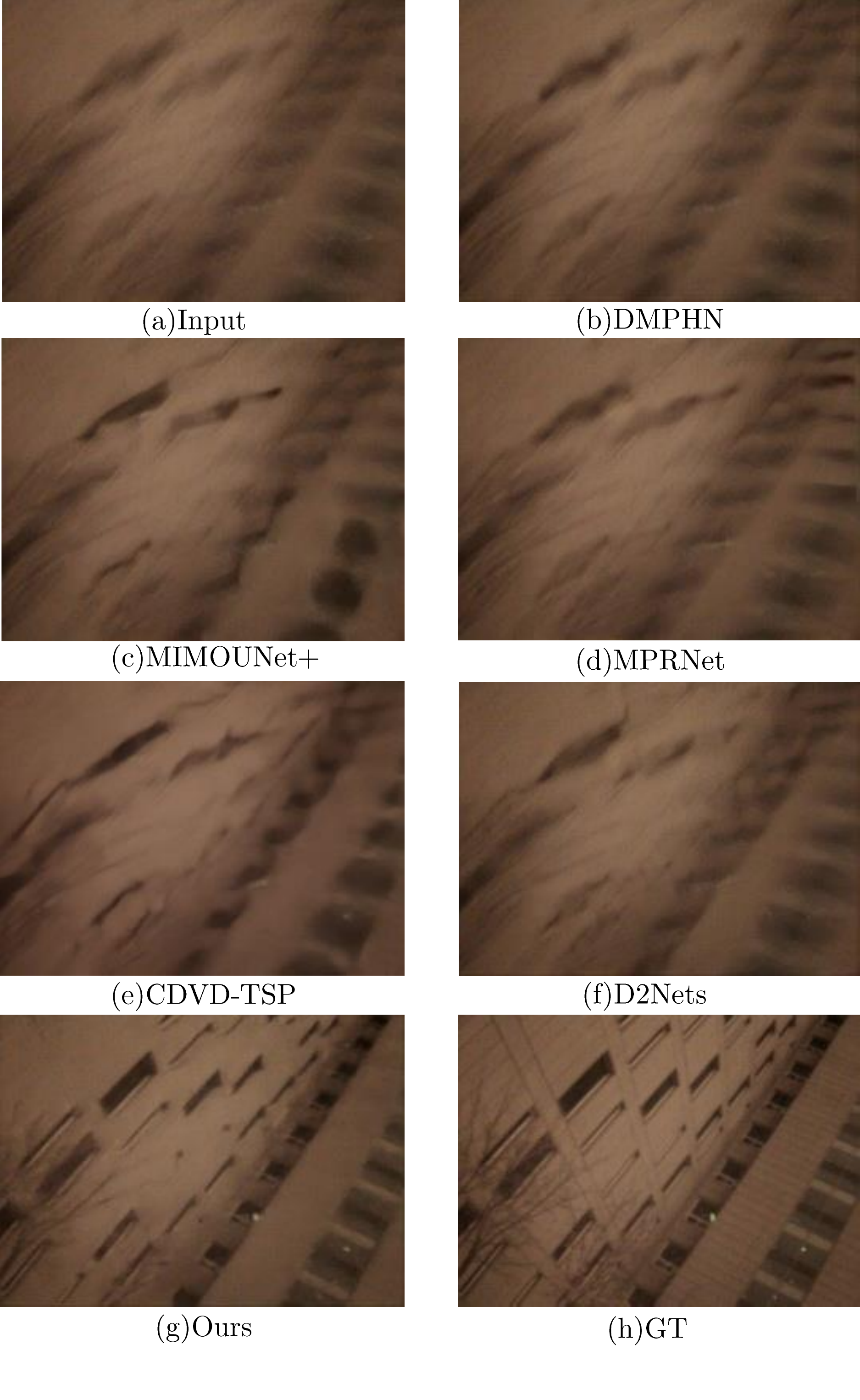}
\caption{Visual comparison of unknown exposure time blurry video frames on our real-world event datasets.}
\label{fig:color_dvs_4}
\end{figure}

\newpage
\begin{figure}[!b]
\centering
\includegraphics[width=0.9\columnwidth]{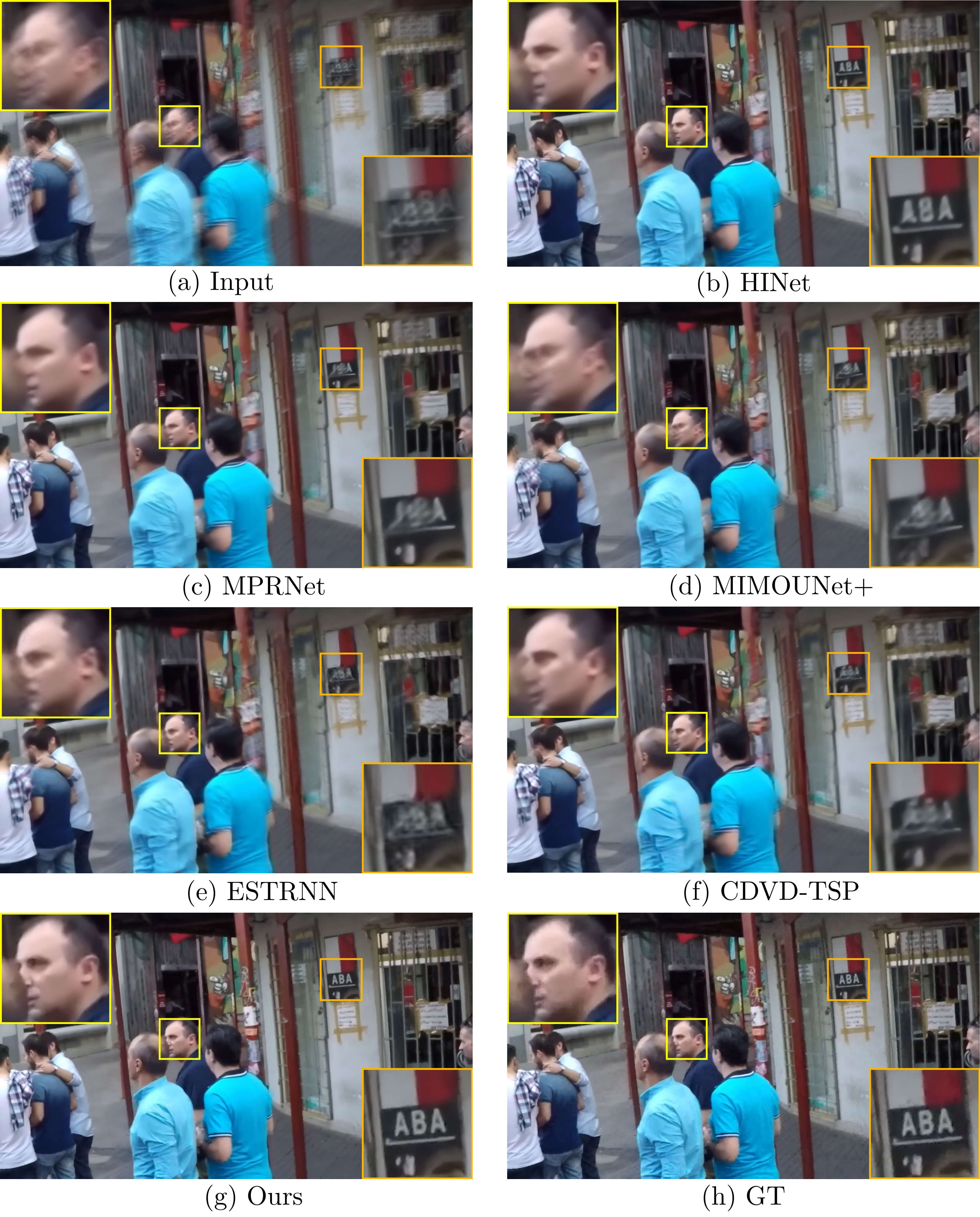}
\caption{Visual comparison on the GoPro-15fps datasets.}
\label{fig:GOPRO_1}
\end{figure}

\newpage
\begin{figure}[!b]
\centering
\includegraphics[width=0.9\columnwidth]{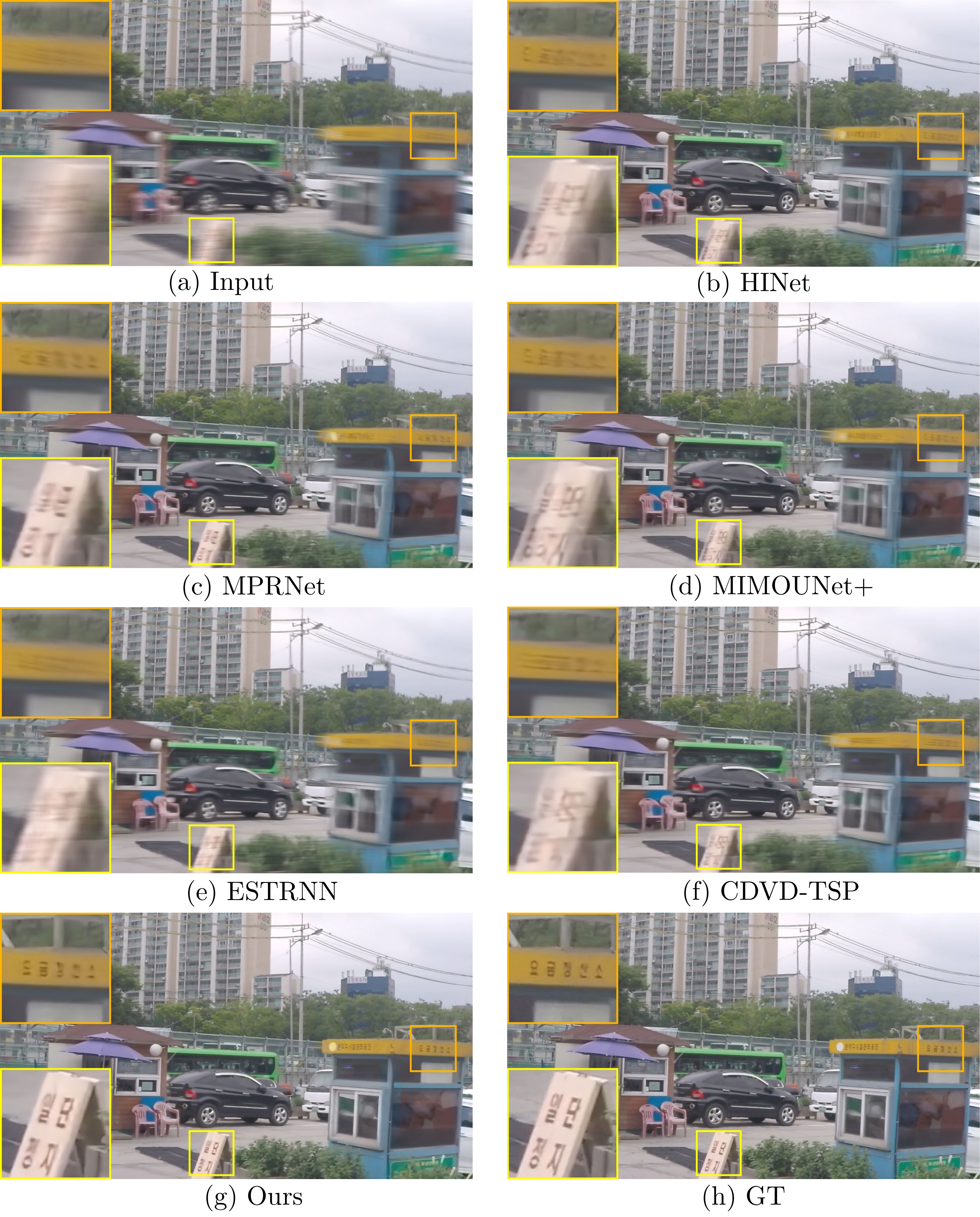}
\caption{Visual comparison on the GoPro-15fps datasets.}
\label{fig:GORPO_2}
\end{figure}

\newpage
\begin{figure}[!b]
\centering
\includegraphics[width=0.9\columnwidth]{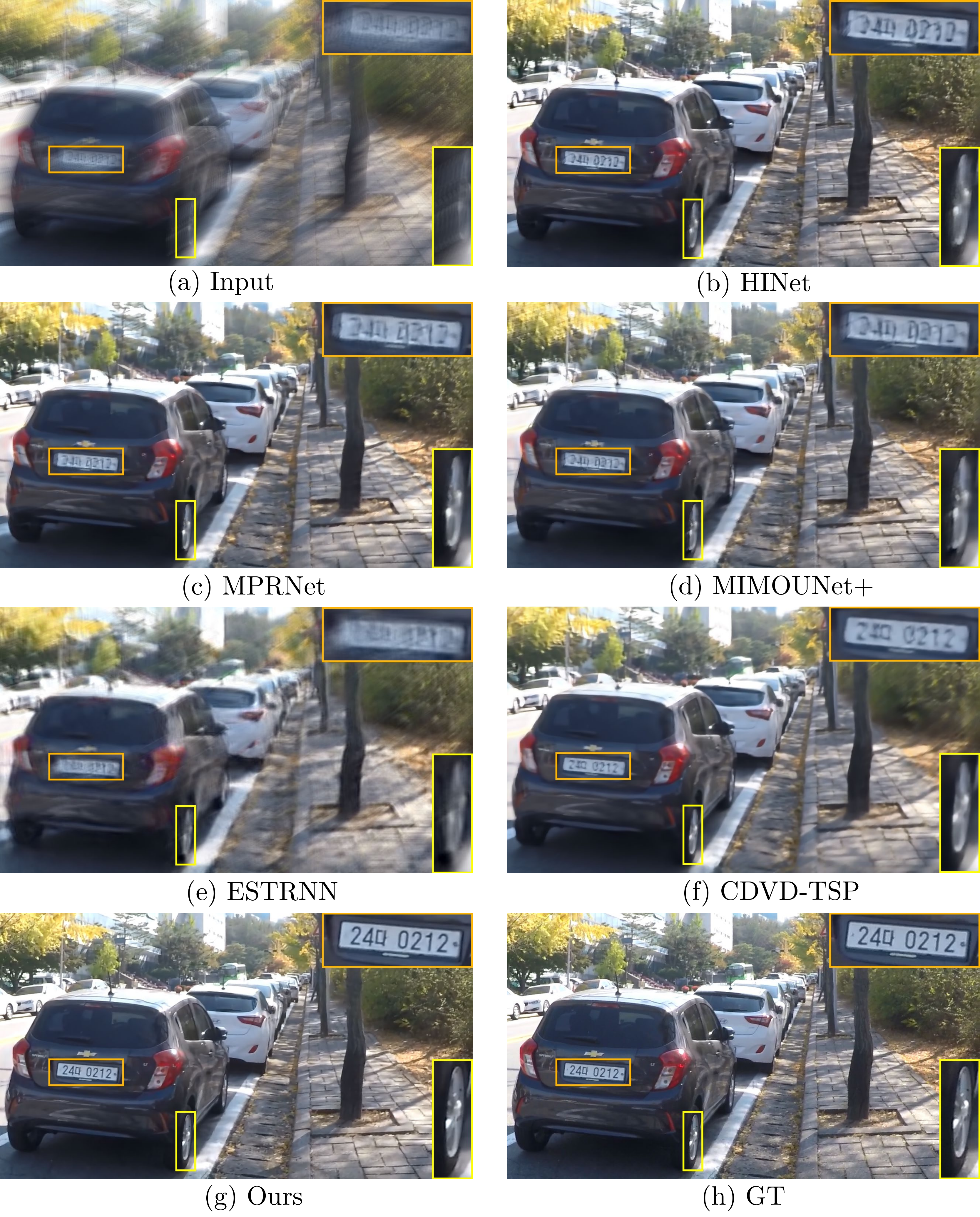}
\caption{Visual comparison on the GoPro-15fps datasets.}
\label{fig:GORPO_3}
\end{figure}

\newpage
\begin{figure}[!b]
\centering
\includegraphics[width=0.9\columnwidth]{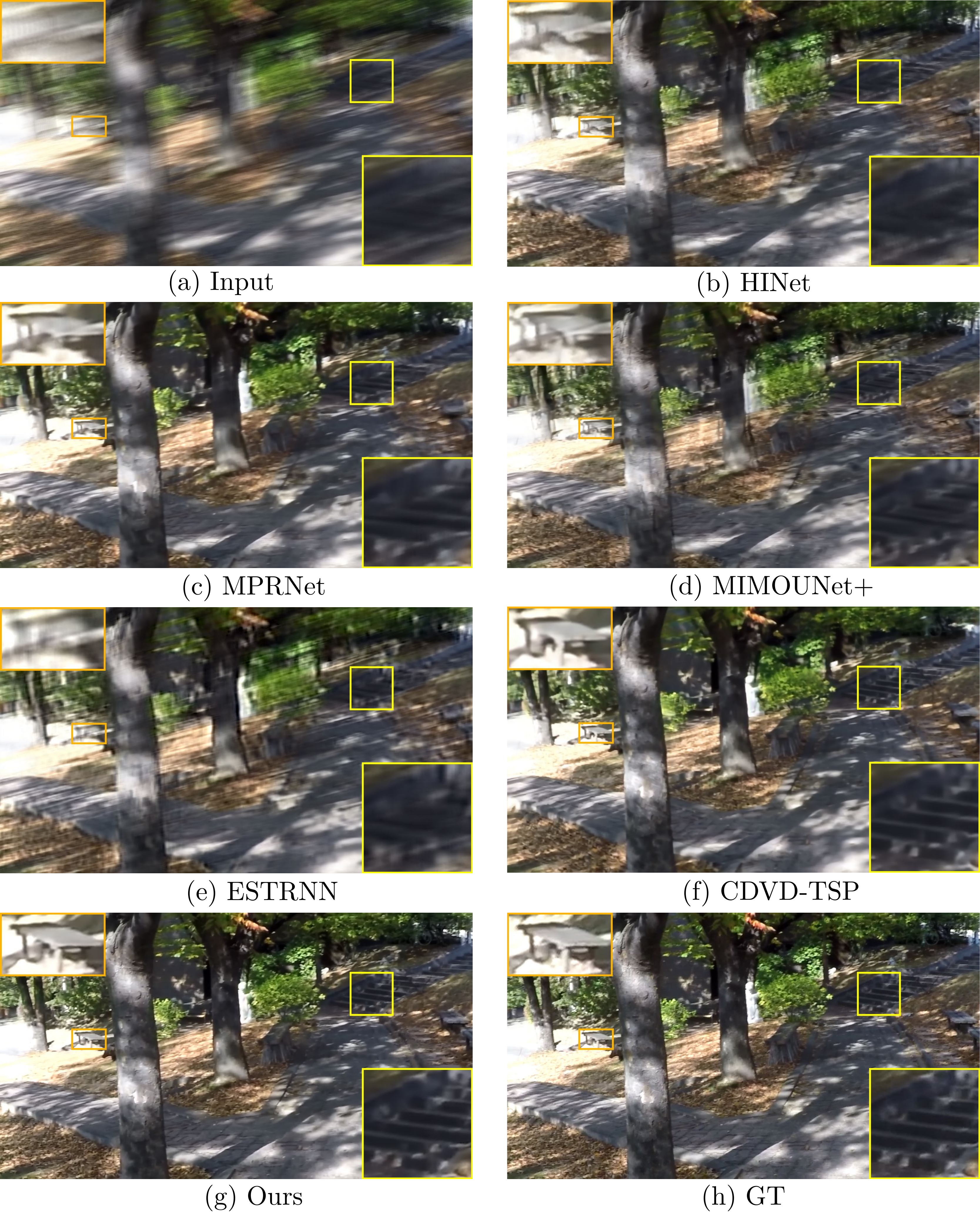}
\caption{Visual comparison on the GoPro-15fps datasets.}
\label{fig:GORPO_4}
\end{figure}

\clearpage
%
%
\end{document}